\documentclass{mc_nankai}
\usepackage[utf8]{inputenc} 
\usepackage[T1]{fontenc}    
\usepackage{url}            
\usepackage{booktabs}       
\usepackage{amsfonts}       
\usepackage{nicefrac}       
\usepackage{microtype}      
\usepackage{multirow}
\usepackage{array}
\usepackage{amssymb}
\usepackage{booktabs}
\usepackage{amsmath}
\usepackage{adjustbox}
\usepackage{amssymb}

\usepackage{subcaption}
\usepackage{graphicx}
\usepackage{caption}
\usepackage{arydshln}
\usepackage{tikz}
\usepackage{wrapfig}
\usepackage{float}
\usepackage{multirow}
\usepackage{pifont}
\usepackage{array}
\usepackage{soul}
\usepackage{makecell}
\newcommand{\cmark}{\ding{51}}%

\def\MyMthd{DepthAnything-AC}

\newcommand{\tablestyle}[2]{\setlength{\tabcolsep}{#1}\renewcommand{\arraystretch}{#2}\centering\small}

\title{Depth Anything at Any Condition}

\author[1]{Boyuan Sun$^{*}$}
\author[1]{Modi Jin$^{*}$}
\author[1]{Bowen Yin}
\author[1]{Qibin Hou$^{\dagger}$}

\affiliation[1]{VCIP, School of Computer Science, Nankai University}

\affiliation[]{$^{*}$Equal Contribution.  $^{\dagger}$Corresponding author.}

\abstract{
We present Depth Anything at Any Condition (DepthAnything-AC), a foundation monocular depth estimation (MDE) model capable of handling diverse environmental conditions. 
Previous foundation MDE models 
achieve impressive performance across general scenes but not perform well in complex open-world environments that involve challenging conditions, such as illumination variations, adverse weather, and sensor-induced distortions.
To overcome the challenges of data scarcity and the inability of generating high-quality pseudo-labels from corrupted images, we propose an unsupervised consistency regularization finetuning paradigm that requires only a relatively small amount of unlabeled data. 
Furthermore, we propose the Spatial Distance Constraint to explicitly enforce the model to learn patch-level relative relationships, resulting in clearer semantic boundaries and more accurate details.
Experimental results demonstrate the zero-shot capabilities of DepthAnything-AC across diverse benchmarks, including real-world adverse weather benchmarks, synthetic corruption benchmarks, and general benchmarks. 
}

\mclink[Project Page]{\url{https://ghost233lism.github.io/depthanything-AC-page}}
\mclink[Code]{\url{https://github.com/HVision-NKU/DepthAnythingAC }}

\begin{document}

\maketitle
\justifying

\section{Introduction} \label{sec:intro}

Monocular depth estimation (MDE)~\cite{Wei2021CVPR, godard2019digging, godard2017unsupervised,bian2021tpami} is a crucial task in computer vision as it enables the direct extraction of a depth map from a single image. Recently, foundation MDE models~\cite{lin2024promptda, hu2024metric3dv2, ke2023repurposing,li2023patchfusion, song2025depthmaster, he2025distill}, especially Depth Anything series~\cite{yang2024depth, yang2024depthv2} and DepthPro~\cite{bochkovskii2024depthpro}, have emerged, demonstrating impressive zero-shot capabilities in complex and diverse scenarios.
These models significantly reduce the cost of acquiring accurate depth maps in general scenarios. As a result, depth maps have not only propelled advancements in traditional perception-driven tasks, such as RGB-D segmentation~\cite{zhang2019pattern,yin2024dformer, dformerv2}, robotics~\cite{li2024manipllm,li2024visual}, autonomous driving~\cite{hu2024drivingworld, wang2021depth,sc_depthv3}, and 3D reconstruction~\cite{yin2022towards, kerbl20233d, msnerf_2025_TPAMI, wang2025vggt, deng2024boost}, but have also been increasingly adopted in emerging domains like AI-generated content (AIGC)~\cite{zhang2023adding, shriram2024realmdreamer} and large multi-modal models~\cite{cai2024spatialbot, pang2024depthhelpsimprovingpretrained, zhao2025llava,sun2025llava} to enhance scene understanding capabilities.

\begin{figure}[t] 
\vspace{-5mm}
  \centering
  \includegraphics[width=1.0\textwidth]{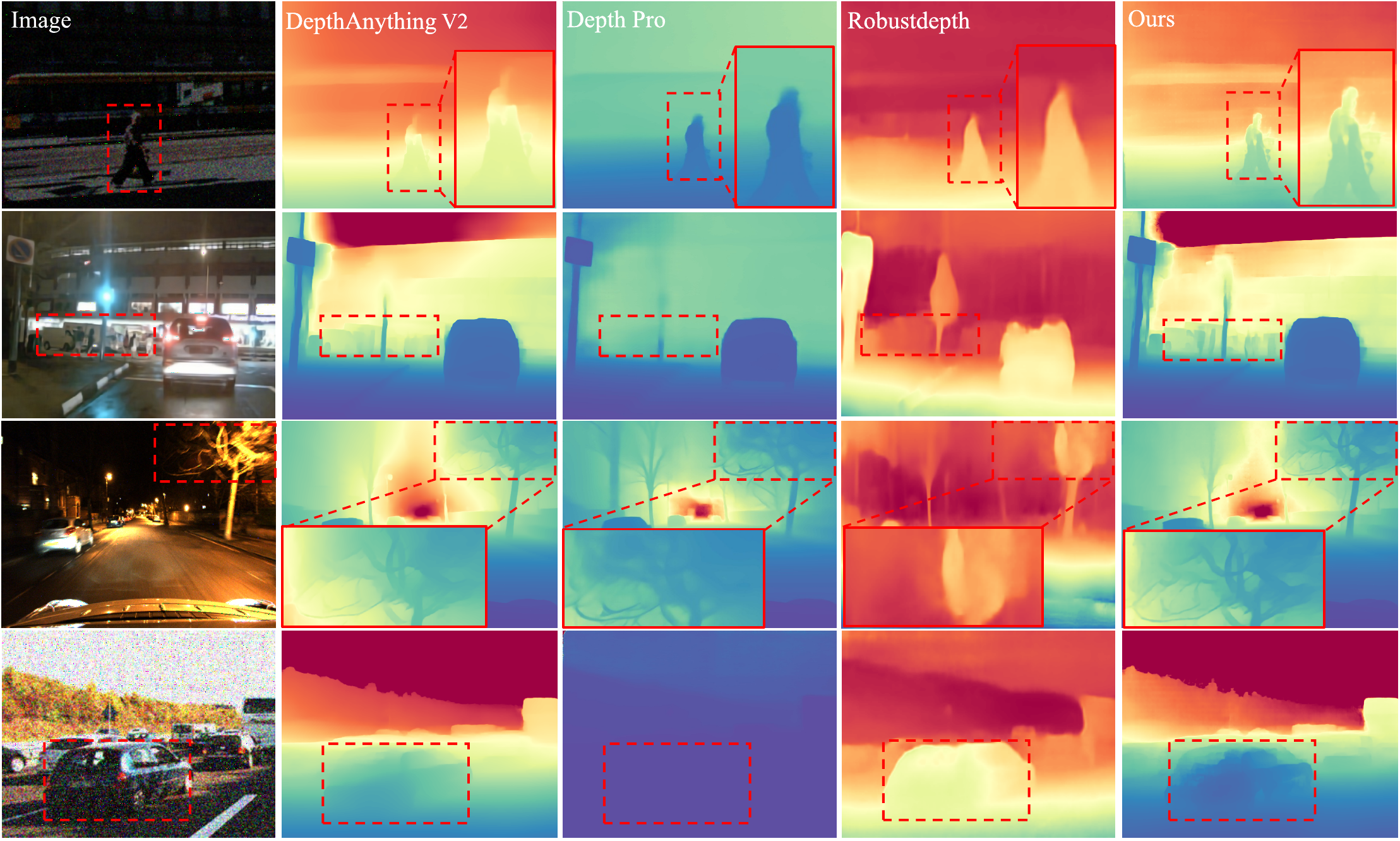} 
  \caption{\textbf{ Zero-shot predictions.} 
  \MyMthd{} achieves better details compared to other approaches, including Depth Anything V2~\cite{yang2024depthv2}, DepthPro~\cite{bochkovskii2024depthpro}, and RobustDepth~\cite{saunders2023robust}, under challenging lighting and climatic conditions.
  } 
  \label{fig:teaser} 
\end{figure}

Compared to traditional approaches~\cite{fu2018deep, eigen2014depth}, recent foundation MDE models often utilize a massive amount of training data to achieve robust performance in general scenarios. However, since the main training data~\cite{kitti, cabon2020vkitti2, cho2021diml} are captured under normal lighting and weather conditions, these models often struggle to perform well in more challenging environments, particularly those involving complex illumination, diverse weather patterns, and sensor-induced distortions, as illustrated in \figref{fig:teaser}. 
This limitation restricts the applicability of foundation MDE models in open-world scenarios.

In this paper, we aim to propose a foundation MDE model capable of extracting depth maps under complex conditions while maintaining its general abilities.
The main challenge is the lack of image-depth pairs for complex weather scenarios. 
While some existing datasets~\cite{caesar2020nuscenes, kong2023robodepth, robotcar} attempt to include complex weather scenarios, they often focus on a specific domain, which poses the potential risk of undermining the foundation model if we directly finetune on them. 
Meanwhile, as shown in \figref{fig:teaser}, existing approaches~\cite{yang2024depthv2, bochkovskii2024depthpro} struggle in complex real-world corrupted scenarios, making it difficult to obtain reliable pseudo-labels directly from such data.
%
To address these issues, we propose a consistency regularization paradigm based on perturbation. 
%
Specifically, this paradigm utilizes a relatively small amount of unlabeled data from general scenes and applies augmentations such as lighting, blur, weather, and contrast on them, encouraging the model to predict consistent results among different views.  This aims to equip the model with robustness to complex lighting and weather conditions while maintaining its performance in general scenes.

Furthermore, existing models~\cite{midas,bhat2023zoedepth, yang2024depth} focus solely on per-pixel depth differences while neglecting the relative positional relationships between pixels, which limits their ability to perceive object structures and makes them more susceptible to disruptions at object boundaries. 
The unsatisfactory results of DepthAnything V2~\cite{yang2024depthv2} and DepthPro~\cite{bochkovskii2024depthpro} on object boundaries and details in \figref{fig:teaser} further support this point.
To address this, we attempt to derive some insights from the underlying spatial geometric relationships. Specifically, 
we propose combining positional information with depth values to construct a more comprehensive representation for measuring patch-wise spatial distances.
As shown in \figref{fig:prior}, spatial distance maps can emphasize the semantic information of the object at the corresponding position with clear boundaries. 
Based on this, we further propose the Spatial Distance Constraint for monocular relative depth estimation to constrain the relative positioning between patches.
By introducing spatial distance constraints between patches, the model can perceive the positional relationships and shapes of objects. 
This position-aware optimization method not only enhances the accuracy of object boundary localization but also reduces the prediction ambiguity caused by texture loss.

Our contributions are summarized as follows:
\begin{itemize}
    \item We propose a perturbation-based knowledge distillation paradigm that does not require any labeled data, which can expand the capabilities of foundation MDE models on real-world corrupted scenes while preserving their performance in general scenarios.
    \item We show that the relative spatial positioning between patches is highly beneficial for modeling semantic boundaries. By extracting spatial distance as semantic priors from depth maps and spatial positions and using it as the spatial distance constraint, we can better reconstruct object structures and boundaries from corrupted texture information.
    \item We propose the Depth Anything at Any Condition (DepthAnything-AC) model to improve depth estimation in more complex open-world scenarios, particularly under complex lighting and weather conditions. Experiments demonstrate that DepthAnything-AC outperforms state-of-the-art approaches in both real-world and synthetic noise scenarios.
\end{itemize}

\vspace{-10pt}

\begin{figure}[t] 
  \centering
  \includegraphics[width=1.0\textwidth]{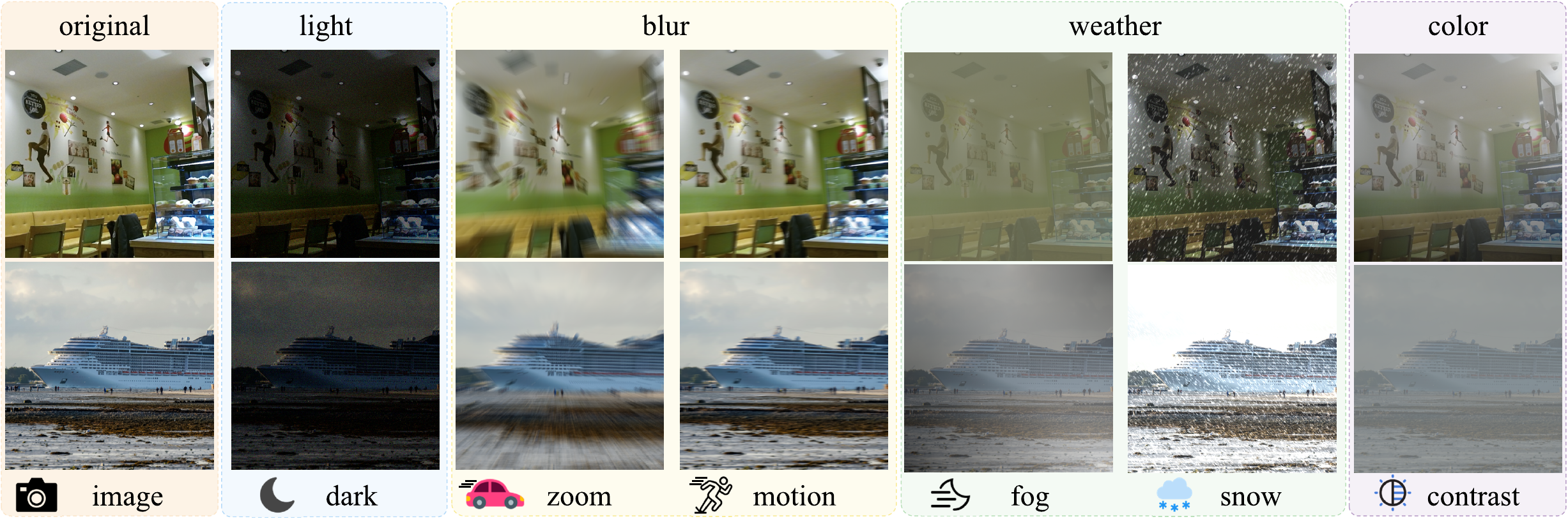} 
  \caption{\textbf{Visualization of perturbation types.} We apply a combination of darkness, weather, blur, and contrast perturbations during training. Note that different weather or blur perturbations do not appear simultaneously. For example, motion blur and zoom blur are mutually exclusive.
} 
  \label{fig:enhance}
\end{figure}

\section{Related Work}
\subsection{Foundation Monocular Depth Estimation Model}
Monocular depth estimation~\cite{bhoi2019monocular, zhao2020monocular, rajapaksha2024deep, wu2022toward, bian2019neurips} is one of the most classic tasks in computer vision, and it has even been studied early before the advent of deep learning~\cite{torralba2002depth, wedel2006realtime, saxena2005learning, liu2010single}.  Although deep learning-based methods~\cite{fu2018deep, eigen2014depth, guo2025murre} have achieved remarkable progress in in-domain scenarios~\cite{bhat2021adabins, kitti, bhat2023zoedepth, wang2025tacodepthefficientradarcameradepth}, their reliance on single-domain data limits their applicability in open-world environments.

MiDaS~\cite{birkl2023midas, Ranftl2022} and Metric3D~\cite{yin2023metric3d, hu2024metric3dv2} are pioneering works toward building a foundation MDE model, which integrates multiple depth datasets.
The Depth Anything series~\cite{yang2024depth, yang2024depthv2} establish a semi-supervised paradigm~\cite{yang2022revisiting, sun2023corrmatch} by leveraging large-scale unlabeled data for training, significantly enhancing the zero-shot performance of MDE models in diverse real-world scenarios.
%
Some approaches tackle the MDE task with generative models~\cite{xu2024depthsplat, saxena2023monocular, fu2024geowizard, he2024lotus, pham2024sharpdepth}. Marigold~\cite{ke2023repurposing} establishes a paradigm based on Stable Diffusion~\cite{rombach2021highresolution} to generate depth maps. DepthLab~\cite{liu2024depthlab} leverages inpainting techniques to enhance structural details in depth predictions. DepthFM~\cite{gui2024depthfm} introduces a flow matching strategy to enable fast and efficient sampling within diffusion-based frameworks.

Although these foundation MDE models have demonstrated impressive zero-shot performance in general ``in-the-wild'' scenarios, they overlook the impact of real-world image degradations, such as various lighting, adverse weather conditions, and sensor-involved distortions, which commonly occur in open-world settings. As a result, their performance may deteriorate under such challenging conditions. Therefore, we aim to propose a more comprehensive foundation MDE model that not only maintains strong general capabilities but also exhibits robustness to more complex scenarios.

\subsection{Robust Depth Estimation Models}
Besides achieving zero-shot capability in general scenarios, obtaining robust monocular depth estimation models remains a critical objective.
Some early self-supervised methods leverage GANs~\cite{CycleGAN2017, goodfellow2020generative} to improve monocular depth estimation in nighttime and adverse weather conditions. Works like ADFA~\cite{ADFA}, ADDS~\cite{ADDS}, and RNW~\cite{RNW} learn features from complex scenes, while ITDFA~\cite{ITDFA} 
generates nighttime pseudo-labels for supervision.
Recently, many works utilize the diffusion-based pipeline~\cite{tosi2024diffusion,mao2024stealing,wang2024digging} to enhance the robustness of depth estimation models. 


To address dynamic ambiguity, methods like DeFeat-Net~\cite{defeat-net} and R4Dyn~\cite{r4} propose joint learning and auxiliary modality integration, while others~\cite{vankadari2022sun,ekf-imu,fan2025depth} focus on illumination variations with targeted constraints. Recently, increasing attention has shifted toward image-level analysis. For example, Robodepth~\cite{kong2023robodepth} presents a depth enhancement scheme and robustness benchmark, while works like Robust-depth~\cite{saunders2023robust}, Weatherdepth~\cite{wang2024weatherdepth}, STEPS~\cite{zheng2023steps}, and Syn2Real~\cite{yan2025synthetic} explore robustness under challenging lighting and weather through data perturbations, augmentations, or joint learning.

However, certain issues have been identified with these works. First, unlike foundation MDE models, they are primarily designed for specific domains such as autonomous driving, which restricts their applicability in open-world scenarios. Moreover, despite employing complex strategies to enhance model robustness, the lack of sufficient semantic perception capabilities often results in suboptimal boundary delineation and insufficient detail recovery.

\begin{figure}[t] 
  \centering
  \includegraphics[width=1.0\textwidth]{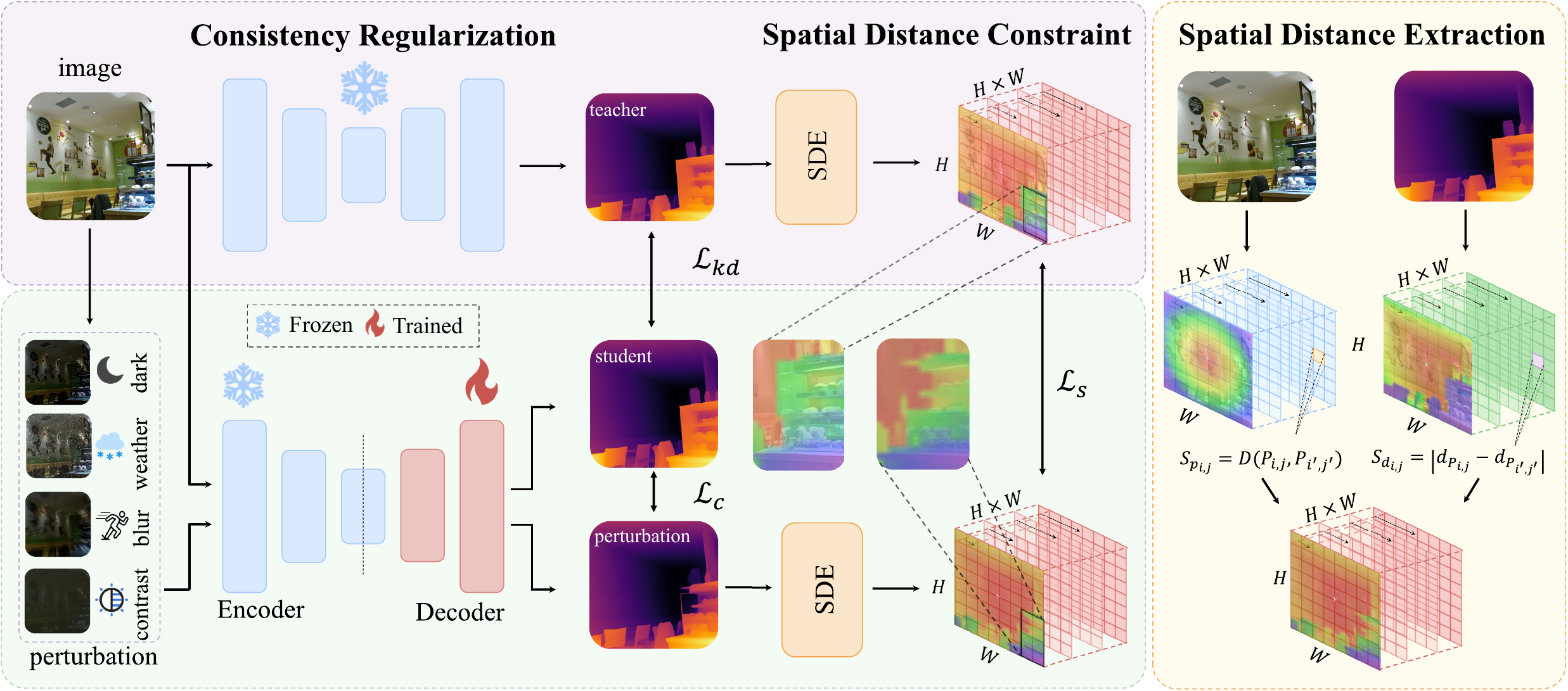} 
  \caption{\textbf{Training pipeline for DepthAnyting-AC.} The perturbation consistency framework encourages DepthAnything-AC to generate consistent predictions under augmentations while retaining generality via the frozen original model. To enhance semantic boundaries and details, the spatial distance constraint is used to strengthen the understanding of inter-patch relationships.
  } 
  \label{fig:pipeline} 
\end{figure}

\section{Depth Anything at Any Condition}

The overall framework of the proposed approach is depicted in \figref{fig:pipeline}.
Our \MyMthd{} consists of a perturbation-based consistency framework that encourages the model to produce consistent outputs for unlabeled images before and after different perturbations, thereby enhancing its robustness in complex scenarios, and a spatial distance constraint that enforces geometric relationships between patches to help recover object boundaries and details from corrupted images.


\subsection{Perturbation-Based Consistency Framework}
\label{consistency framework}
Considering the unsatisfactory performance of existing models on real-world corrupted images and the lack of densely annotated images, we propose a perturbation-based unsupervised consistency framework to extend the capabilities of the existing foundational MDE model in complex scenarios.

We first consider simulating common scenarios that degrade image quality in the real world. Given a normal image from a general scene, we focus on four typical scenarios: lighting, weather, blurriness, and contrast. Specifically, for each scenario, we apply perturbations with varying intensity levels as detailed in \figref{fig:enhance}. For each image, one scenario is selected with a certain probability, and the corresponding perturbation is applied accordingly.
Therefore, for an unlabeled image $x_i$, the image augmented through regular transformations is denoted as $x_i^w$, and the image further processed with perturbations is denoted as $x_i^s = \mathcal{A}(x_i^w)$.

Then, from the perspective of consistency regularization, we design the perturbation-based unsupervised consistency framework, as shown in \figref{fig:pipeline}. Specifically, given a mini-batch of unlabeled images $\{x_i\}$ and a foundation MDE model $\mathcal{F}$, we define the consistency loss $\mathcal{L}_c$ as:
\begin{equation}
\label{consistency}
    \mathcal{L}_c = \frac{1}{N} \sum_{i=1}^{N} \ell(\mathcal{F}(x^w_i), \mathcal{F}(x^s_i)),
\end{equation}
where $x^w_i$ and $x^s_i$ are with regular and perturbed images, respectively.
This loss function enforces consistency between the model's predictions under normal and perturbed scenarios, encouraging the model to provide consistent outputs across different corrupted conditions. 
The $\ell(y, \hat{y})$ in \eqnref{consistency} is a measurement of the prediction disparity $y$ and target disparity $\hat{y}$ (Disparity is the inverse of relative depth, which is the direct output of the foundation MDE model). Inspired by DepthAnything~\cite{yang2024depth}, we define the affine-invariant loss $\ell$ as:
\begin{equation}
    \ell = \frac{1}{HW} \sum_{j=1}^{HW}|\frac{y_j - t(y)}{s(y)} - \frac{y_j - t(\hat{y})}{s(\hat{y})}|,
\end{equation}
where $t(y)$ and $s(y)$ are the factors used to balance the scale between prediction and target.
\begin{equation}
    t(y) = \text{Mean}(y),  s(y) = \frac{1}{HW}\sum_{j=1}^{HW} |y_j - t(y)|.
\end{equation}

Also, while consistency regularization encourages the model to provide consistent predictions across images with different perturbations, the lack of precise label supervision introduces a potential risk: The model might prioritize consistency over correctness, which could degrade its performance in general scenes. To tackle this, we keep the initial foundation MDE model $\hat{\mathcal{F}}$ frozen as guidance and use a knowledge distillation loss to guide the model's predictions in the normal scene, as follows:
\begin{equation}
\label{kd loss}
    \mathcal{L}_{kd} = \frac{1}{N} \sum_{i=1}^{N} \ell(\mathcal{F}(x^w_i), \hat{\mathcal{F}}(x^w_i)).
\end{equation}

\begin{figure}[t] 
  \centering
  \includegraphics[width=1.0\textwidth]{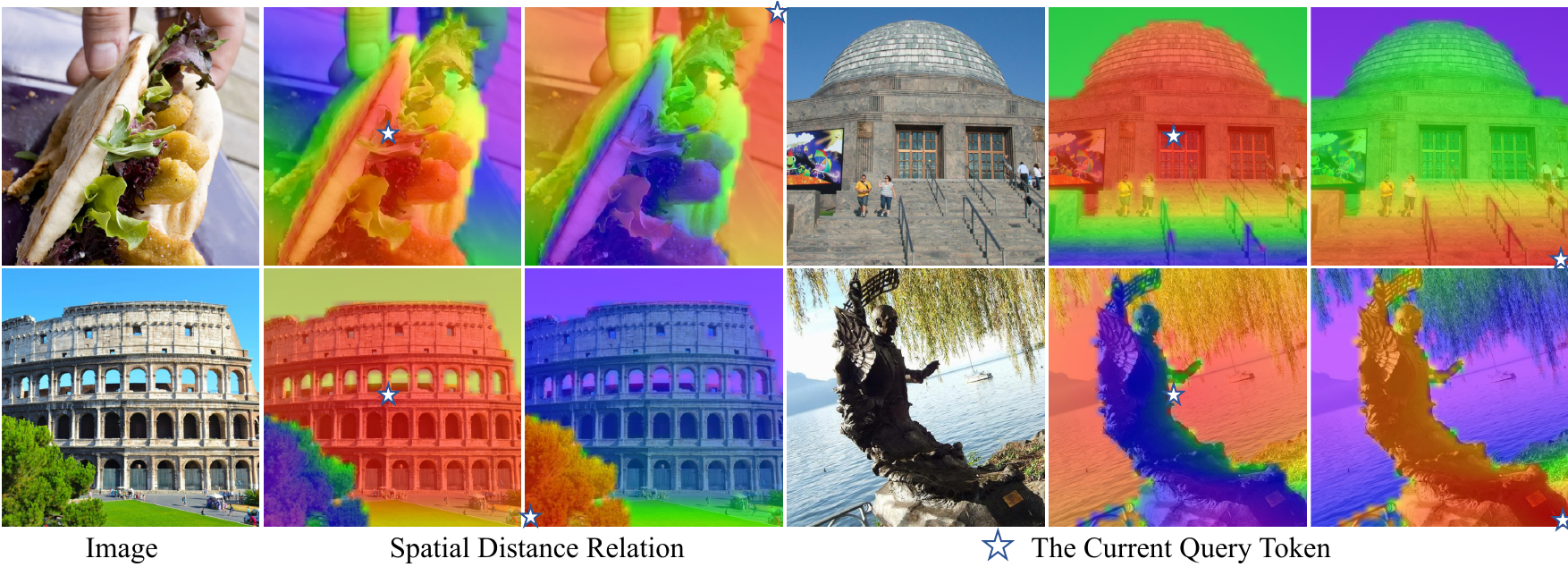} 
  \caption{\textbf{Visualization of spatial distance relationships corresponding to  different query tokens.}  The pentagram denotes the current query token, illustrating the spatial distance between the patch containing the pentagram and all other patches in the frame.} 
  \label{fig:prior} 
\end{figure}

\subsection{Spatial Distance Constraint}
Beyond the perturbation-based consistency framework, real-world corruptions often degrade image quality, making it challenging for existing foundation MDE models to accurately distinguish object boundaries and capture fine-grained details. To address this, we propose the Spatial Distance Relation (SDR), a geometric prior that captures relative distance among objects. It can serve as a complement to enhance the semantic perception capabilities of MDE models under adverse conditions.

The SDR $S_D$ consists of two parts, the position relation $S_p$ and the depth relation $S_d$.
Given an unlabeled image $x_i$ consisting of $H \times W$ patches, we denote the spatial coordinate of each patch as $p_{i,j}$, where $i \in \{1, \dots, H\}$ and $j \in \{1, \dots, W\}$. To characterize the spatial structure, we first compute the position relation $S_{p_{i,j}}$ by calculating the Euclidean distance between each pair of patches:
\begin{equation}
\label{sp}
    S_{p_{i,j}} = \{\left\| p_{i,j} - p_{m,n} \right\|_2\}, 
\end{equation}
where $m\in \{1,2,...,H\}$ and $n\in \{1,2,...,W\}$.
This produces a position relation matrix $S_p \in \mathbb{R}^{HW \times HW}$, which contains pairwise positional distances between each pair of patches. The position relation matrix is subsequently min-max normalized.





For the depth relation $S_{d_{ij}}$ of patch $p_{i,j}$, it is defined as the absolute difference in predicted disparities between patch $p_{i,j}$ and some patch $p_{m,n}$:
\begin{equation}
S_{d_{i,j}} = \{|\mathcal{F}(x_i)_{i,j}- \mathcal{F}(x_i)_{m,n}|\}.
\label{eq:dp}
\end{equation}


Since the position relation $S_p$ encodes the relative positional distance between any two patches in the image plane and the depth relation $S_d$ represents the disparity distance between them, the relative geometric distance between each pair of patches can be formulated as:

\begin{equation}
S_D=\sqrt{S_p^2+S_d^2} \in \mathbb{R}^{HW\times HW}.
\label{eq:final}
\end{equation}

\begin{table}[t]
  \centering
  \small
\setlength{\abovecaptionskip}{2pt}
\tablestyle{18.5pt}{1}
    \begin{tabular}{lcccccc}
    \toprule
    Dataset & Indoor & Outdoor & Samples & Used & Ratio \\
    \midrule
    ADE20k~\cite{zhou2017scene,zhou2019semantic} & \cmark     & \cmark     & 20K & 20K & 100\% \\
    MegaDepth~\cite{MegaDepthLi18} &       & \cmark     & 128K & 100K& 78\%\\
    DIML~\cite{cho2021diml,kim2016structure,cho2021deep,kim2018deep,kim2017deep}  & \cmark     & \cmark  & 927K    & 80K & 8.6\%\\
    VKITTI2~\cite{cabon2020vkitti2} &       & \cmark  & 42K   & 42K & 100\%\\
    HRWSI~\cite{gaidon2016virtual,Xian_2020_CVPR} &       & \cmark  & 20K  & 20K& 100\% \\
    SA-1B~\cite{kirillov2023segment}   & \cmark     & \cmark   & 11.1M  & 140K & 1.3\%\\
    COCO~\cite{lin2014microsoft}  & \cmark     & \cmark     & 120K  &120K & 100\%\\
    Pascal VOC 2012~\cite{everingham2010pascal}   & \cmark     & \cmark  & 10K   & 10K  & 100\%\\
    AODRaw~\cite{li2024aodraw} &   \cmark    & \cmark  & 80K   & 8K & 10\% \\
    \bottomrule
    \end{tabular}
    
    \caption{Datasets used for training. In total, our DepthAnything-AC is fine-tuned on 540K unlabeled images, which is much less than the number used in the DepthAnything series~\cite{yang2024depth, yang2024depthv2}.}
  \label{tab:train data}%
\end{table}
\begin{table}[t]
  \centering
  \small
  \setlength{\abovecaptionskip}{2pt}
  \tablestyle{8pt}{1}
    \begin{tabular}{lcccccc}
    \toprule
    Method & Encoder & \textbf{DA-2K} &\textbf{ DA-2K dark} &\textbf{ DA-2K fog} & \textbf{DA-2K snow} & \textbf{DA-2K blur} \\
    \midrule
     DynaDepth~\cite{ekf-imu}&   ResNet     & 0.655      &     0.652  &     0.613  &  0.605     & 0.633\\
          EC-Depth~\cite{zhu2023ecdepth}&   ViT-S     & 0.753      &     0.732  &     0.724  &  0.713     & 0.701 \\
          STEPS~\cite{zheng2023steps}& ResNet&  0.577     &     0.587  &    0.581   &    0.561   &   0.577     \\
        robustdepth~\cite{saunders2023robust}& ViT-S        & 0.724      &     0.716  &    0686   &   0.668    & 0.680 \\
          weather-depth~\cite{wang2024weatherdepth}& ViT-S        & 0.745      &     0.724  &    0.716   &   0.697    & 0.666 \\
          DepthPro\cite{bochkovskii2024depthpro} &  ViT-S      &   0.947    &   0.872    &    0.902   &   0.793    & 0.772 \\
          DepthAnything V1~\cite{yang2024depth}  & ViT-S      &    0.884  &     0.859  &    0.836   &     \underline{0.880}  &0.821  \\
        DepthAnything V2~\cite{yang2024depthv2} & ViT-S      &   \underline{0.952}   &    \underline{0.910}   &    \underline{0.922}  &     \underline{0.880} &\underline{0.862}  \\
      \textbf{\MyMthd{}}    &   ViT-S    & \textbf{0.953}     &   \textbf{0.923}    &   \textbf{0.929}    &  \textbf{0.892}    &  \textbf{0.880} \\
    \bottomrule
    \end{tabular}%
  \caption{Quantitative results on the enhanced multi-condition DA-2K benchmark, including complex light and climate conditions. 
  The first and second ranked performances are indicated by \textbf{bold} and \underline{underline}. 
  The evaluation metric used is \textbf{Accuracy} $\uparrow$.}
  \label{tab:DA-2K}%
\end{table}

The physical interpretation of the spatial distance relation $S_D$ is a geometric distance between any pair of patches. Each row of $S_D$ reflects the degree of geometric proximity to a specific patch, inherently encoding rich semantic information.
As illustrated in \figref{fig:prior}, the constructed spatial relations closely correlate with the semantic structure of objects within the image. This enables the model to leverage semantic priors in a training-free manner. Based on SDR, we propose the Spatial Distance Relation loss function as a constraint:
\begin{equation}
\mathcal{L}_{s} = \frac{1}{N} \sum_{i=1}^{N} (S_{D}(\mathcal{F}(x^s_i)) - \hat{S}_{D}(\hat{\mathcal{F}}(x^w_i)))^2.
\label{eq:spatial loss}
\end{equation}

This loss function encourages the model to maintain consistency between the spatial distance relations derived from the perturbed image and those from the original image predicted by the original model. It facilitates the recovery of semantic boundaries and fine details degraded by real-world corruptions.

Equipped with the perturbation-based consistency framework and the spatial distance constraint, the overall supervision is a combination of $\mathcal{L}_{c}$,  $\mathcal{L}_{kd}$, and  $\mathcal{L}_s$:
\begin{equation}
    \mathcal{L} = \lambda_1\mathcal{L}_{c} + \lambda_2\mathcal{L}_{kd} + \lambda_3\mathcal{L}_s,
\end{equation}
where $\lambda_1$, $\lambda_2$, and $\lambda_3$ are weighting coefficients and set to [1/3, 1/3, 1/3] by default.

\section{Experiments}

\subsection{Experiment Setup}
\label{sec: setup}
\myPara{Implementation details.} 
DepthAnything-AC is fine-tuned based on DepthAnythingV2~\cite{yang2024depthv2}. Specifically, it adopts ViT-S~\cite{dosovitskiy2020vit} as the backbone and DPT~\cite{midas} as the decoder. During training, we use the AdamW~\cite{loshchilov2017decoupled} optimizer with an initial learning rate of 5e-6, a weight decay of 0.01, a crop size of 518×518, and a batch size of 16. We train the model for 20 epochs with 4 NVIDIA RTX 3090 GPUs.

\myPara{Training datasets.} The datasets we collect for fine-tuning \MyMthd{} model are summarized in \tabref{tab:train data}.
Our training set consists of 540K samples in total, which is relatively small compared to the Depth Anything series (around 63M)~\cite{yang2024depth, yang2024depthv2}, making up less than 1\%. Among these datasets, except for AODRaw~\cite{li2024aodraw}, which includes some scenes under complex natural conditions, the rest consist of images captured under normal lighting and weather conditions.



\textbf{Evaluation benchmarks.}
We evaluate the zero-shot performance of \MyMthd{} across various benchmarks. Specifically, we test on the enhanced multi-weather DA-2K benchmark, real-world scenarios with complex lighting and weather conditions, and synthetic benchmarks. Additionally, we assess performance on general benchmarks to ensure that \MyMthd{} maintains its generalization ability.
Specifically, we select the NuScenes-night~\cite{caesar2020nuscenes}, Robotcar-night~\cite{robotcar}, and Driving-Stereo~\cite{yang2019drivingstereo} as the real-world benchmarks, and the KITTI-C~\cite{kong2023robodepth} as the synthetic benchmark. As for general benchmarks, following previous methods, we use KITTI~\cite{kitti}, NYU-D~\cite{nyu}, Sintel~\cite{sintel}, ETH3D~\cite{eth3d}, and DIODE~\cite{diode_dataset}. 
All experiments are performed on a single NVIDIA RTX 3090 GPU.

\begin{table}[t]
  \small
  \centering
  \setlength{\abovecaptionskip}{2pt}
  \tablestyle{2.5pt}{1.0}
  \begin{tabular}{lccccccccccccccc}  
    \toprule
    \multirow{2}{*}{Method} & \multirow{2}{*}{Encoder} & 
    \multicolumn{2}{c}{\textbf{NuScenes-night}} & 
    \multicolumn{2}{c}{\textbf{RobotCar-night}} & 
    \multicolumn{2}{c}{\textbf{DS-rain}} & 
    \multicolumn{2}{c}{\textbf{DS-cloud}} & 
    \multicolumn{2}{c}{\textbf{DS-fog}} \\
    \cmidrule(l{2pt}r{2pt}){3-4} 
    \cmidrule(l{2pt}r{2pt}){5-6} 
    \cmidrule(l{2pt}r{2pt}){7-8} 
    \cmidrule(l{2pt}r{2pt}){9-10} 
    \cmidrule(l{2pt}r{2pt}){11-12}
    & & AbsRel$\downarrow$ & $\delta$1$\uparrow$ & AbsRel$\downarrow$ & $\delta$1$\uparrow$ & AbsRel$\downarrow$ & $\delta$1$\uparrow$ & AbsRel$\downarrow$ & $\delta$1$\uparrow$ & AbsRel$\downarrow$ & $\delta$1$\uparrow$ \\
    \midrule
   
    DynaDepth~\cite{ekf-imu} & ResNet & 0.381&0.394 &0.512  &0.294  & 0.239&0.606  &0.172  &0.608  &0.144 & 0.901 \\
    EC-Depth~\cite{zhu2023ecdepth}& ViT-S& 0.243& 0.623  & \underline{0.228} & \underline{0.552} &0.155  &0.766  & 0.158 & 0.767  & 0.109 &0.861  \\
    STEPS~\cite{zheng2023steps} & ResNet & 0.252 &  0.588 &0.350  &0.367 & 0.301 & 0.480 &0.252  &0.588  & 0.216 & 0.641 \\
    
    robustdepth~\cite{saunders2023robust} & ViT-S& 0.260 &0.597   & 0.311 & 0.521 & 0.167 & 0.755 & 0.168 & 0.775 & 0.105 & 0.882\\
    weather-depth~\cite{wang2024weatherdepth} & ViT-S & - & - & - & - & 0.158 & 0.764 & 0.160 & 0.767 & 0.105 & 0.879 \\
    Syn2Real~\cite{yan2025synthetic} & ViT-S & - & - &-  &-  & 0.171 & 0.729 & - & - & 0.128 & 0.845 \\
    DepthPro~\cite{bochkovskii2024depthpro} & ViT-S & 0.218 & 0.669 & 0.237 & 0.534& \textbf{0.124} & \textbf{0.841} & 0.158 & 0.779 &\textbf{ 0.102} & \textbf{0.892} \\
    DepthAnything V1~\cite{yang2024depth}  & ViT-S & 0.232 & 0.679 & 0.239 & 0.518 & 0.133 & 0.819 & \underline{0.150} & \textbf{0.801} & \underline{0.098} & \underline{0.891} \\
    DepthAnything V2~\cite{yang2024depthv2}  & ViT-S & \underline{0.200} & \underline{0.725} & 0.239 & 0.518 & \underline{0.125} & \underline{0.840} & 0.151 & \underline{0.798} & 0.103 & 0.890 \\
    \textbf{DepthAnything-AC} & ViT-S & \textbf{0.198} & \textbf{0.727} & \textbf{0.227} &\textbf{0.555} & \underline{0.125} & \underline{0.840} & \textbf{0.149} & \textbf{0.801} & 0.103 & 0.889 \\
    \bottomrule
  \end{tabular}
  \caption{Zero-shot relative depth estimation results on real complex benchmarks.}
  \label{tab:real low quanlity}
\end{table}

\begin{table}[t]
  \small
  \centering
  \setlength{\abovecaptionskip}{2pt}
  \tablestyle{5.8pt}{1.0}
  \begin{tabular}{lccccccccccc}  
    \toprule
    \multirow{2}{*}{Method} & \multirow{2}{*}{Encoder} & 
    \multicolumn{2}{c}{\textbf{Dark}} & 
    \multicolumn{2}{c}{\textbf{Snow}} & 
    \multicolumn{2}{c}{\textbf{Motion}} & 
    \multicolumn{2}{c}{\textbf{Gaussian}} \\
    \cmidrule(l{2pt}r{2pt}){3-4} 
    \cmidrule(l{2pt}r{2pt}){5-6} 
    \cmidrule(l{2pt}r{2pt}){7-8} 
    \cmidrule(l{2pt}r{2pt}){9-10}
    & & AbsRel$\downarrow$ & $\delta$1$\uparrow$ & AbsRel$\downarrow$ & $\delta$1$\uparrow$ & AbsRel$\downarrow$ & $\delta$1$\uparrow$ & AbsRel$\downarrow$ & $\delta$1$\uparrow$ \\
    \midrule

    DynaDepth~\cite{ekf-imu} & ResNet &0.163&0.752 &0.338  &0.393  &0.234  &0.609  & 0.274&0.501   \\
    STEPS~\cite{zheng2023steps} & ResNet & 0.230 & 0.631 &   0.242 & 0.622 & 0.291 & 0.508 & 0.204 & 0.692 \\
    DepthPro~\cite{bochkovskii2024depthpro} & ViT-S & \underline{0.145} & 0.793  & 0.197 & 0.685 & 0.170 & 0.746 & 0.170 & 0.745\\
    DepthAnything V2~\cite{yang2024depthv2} & ViT-S & \textbf{0.130} & \underline{0.832}  & \underline{0.115} & \underline{0.872} & \underline{0.127} & \underline{0.840 }& \underline{0.157} & \underline{0.785} \\
    \textbf{DepthAnything-AC} & ViT-S & \textbf{0.130} & \textbf{0.834}  & \textbf{0.114} &\textbf{0.873} & \textbf{0.126} & \textbf{0.841} & \textbf{0.153} & \textbf{0.793} \\
    \bottomrule
  \end{tabular}
  \caption{Zero-shot relative depth estimation results on synthetic KITTI-C benchmarks. }
  \label{tab:synthetic low quanlity}
\end{table}

\subsection{Zero-Shot Relative Depth Estimation}
\label{sec: performance}
\textbf{Performance on multi-condition DA-2K benchmark.}
DA-2K is a large-scale and high-resolution modern benchmark introduced in DepthAnything v2~\cite{yang2024depthv2}, which evaluates depth models based on pairwise relative depth comparisons.
Beyond that, we further built the multi-condition DA-2K benchmark to evaluate the robust capability of foundation MDE models. 
Specifically, we augment the original DA-2K images with various types of perturbations, namely dark, fog, snow and blur.
As shown in \tabref{tab:DA-2K},  \MyMthd{} outperforms other models under complex lighting and climatic conditions (e.g., 0.862 $\rightarrow$ 0.880 on DA-2K blur).

\textbf{Performance on  real complex lighting and weather scenarios.}
We compare our \MyMthd{} with others on five real complex benchmarks with realistic lighting and climate conditions. As shown in the \tabref{tab:real low quanlity}, our results outperform both robust depth estimation models and foundation MDE models. For instance, compared to Depth Anything V2~\cite{yang2024depthv2}, our \MyMthd{} achieves 0.037 improvements of $\delta1$ metric on the Robotcar-night~\cite{robotcar} benchmark.

\begin{table}[t]

  \centering
  \small
  \setlength{\abovecaptionskip}{2pt}
  \tablestyle{2.5pt}{1.0}
  \begin{tabular}{lccccccccccccccc} 
    \toprule
    \multirow{2}{*}{Method} & \multirow{2}{*}{Encoder} & 
    \multicolumn{2}{c}{\textbf{KITTI}} & 
    \multicolumn{2}{c}{\textbf{NYU-D}} &
    \multicolumn{2}{c}{\textbf{Sintel}} & 
    \multicolumn{2}{c}{\textbf{ETH3D}} & 
    \multicolumn{2}{c}{\textbf{DIODE}} \\
    \cmidrule(l{2pt}r{2pt}){3-4} 
    \cmidrule(l{2pt}r{2pt}){5-6} 
    \cmidrule(l{2pt}r{2pt}){7-8} 
    \cmidrule(l{2pt}r{2pt}){9-10} 
    \cmidrule(l{2pt}r{2pt}){11-12}
    & & AbsRel$\downarrow$ & $\delta$1$\uparrow$ & AbsRel$\downarrow$ & $\delta$1$\uparrow$ & AbsRel$\downarrow$ & $\delta$1$\uparrow$& AbsRel$\downarrow$ & $\delta$1$\uparrow$& AbsRel$\downarrow$ & $\delta$1$\uparrow$\\
    \midrule
    DynaDepth~\cite{ekf-imu} &ViT-S &  0.085 & 0.916 & 0.212 & 0.630 &0.458 &0.391& 0.192 & 0.703 & 0.180& 0.737 \\
    EC-Depth~\cite{zhu2023ecdepth} &ViT-S &  0.075 & 0.931 & 0.220 & 0.606 &0.447  &0.405& 0.181 & 0.779 & 0.172& 0.760 \\
    STEPS~\cite{zheng2023steps} & ResNet &  0.204 & 0.683 & 0.220 & 0.606&0.458  &0.375 & 0.197 & 0.699 & 0.214 & 0.652 \\
    robustdepth~\cite{saunders2023robust} & ViT-S&0.076 &0.929   & 0.174 & 0.718 & 0.411 & 0.436 & 0.185 & 0.776 & 0.182& 0.741\\
    weather-depth~\cite{wang2024weatherdepth} & ViT-S &  0.075 & 0.932 & 0.167 & 0.733 &0.419  &0.443 & 0.181 & 0.771 & 0.170 & 0.760 \\
    DepthPro~\cite{bochkovskii2024depthpro} & ViT-S & \textbf{0.068} & \textbf{0.947} & \underline{0.052} & 0.972 & 0.240 & 0.682& 0.057 & 0.963 & \textbf{0.060} & \textbf{0.952} \\
    DepthAnything V1~\cite{yang2024depth} & ViT-S & \underline{0.082} & \underline{0.936} & 0.053 & \underline{0.972} & \textbf{0.211} & \textbf{0.736} &  \textbf{0.064} & 0.966 & 0.078 & 0.940 \\
    DepthAnything V2~\cite{yang2024depthv2} & ViT-S & 0.083 & 0.934 & \textbf{0.051} & \textbf{0.973} & 0.240 & 0.712 & \textbf{0.064} & \textbf{0.969} & 0.073 & \underline {0.943} \\
    \textbf{\MyMthd{}} & \textbf{ViT-S} & 0.083& 0.934&\textbf{0.051} & \textbf{0.973} & \underline{0.235} & \underline{0.715} & \underline{0.065} & \underline{0.968} & \underline{0.073} & 0.941 \\
    \bottomrule
  \end{tabular}
  \caption{Zero-shot relative depth estimation results on general benchmarks. }
  \label{tab:general_benchmark_results}
\end{table}

\begin{table}[t]

    \centering
    \small
    \setlength{\abovecaptionskip}{2pt}
    \tablestyle{5pt}{1.0}
    \begin{tabular}{cccccccccccccc}
        \toprule
        \multirow{2}{*}{ $\mathcal{L}_{c}$} & \multirow{2}{*}{ $\mathcal{L}_{kd}$} & \multirow{2}{*}{$\mathcal{P}$}  & \multirow{2}{*}{$\mathcal{L}_{s}$} & \multicolumn{2}{c}{\textbf{DA-2K dark}} & \multicolumn{2}{c}{\textbf{NuScenes-night}} & \multicolumn{2}{c}{\textbf{Robotcar-night}} &   \multicolumn{2}{c}{\textbf{Motion}}& \multicolumn{2}{c}{\textbf{Gaussian}}\\
        \cmidrule(lr){5-6} \cmidrule(lr){7-8} \cmidrule(lr){9-10} \cmidrule(lr){11-12}  \cmidrule(lr){13-14}
        & & & &  \multicolumn{2}{c}{Acc$\uparrow$} & AbsRel$\downarrow$ & $\delta1$$\uparrow$ & AbsRel$\downarrow$ & $\delta1$$\uparrow$& AbsRel$\downarrow$ & $\delta1$$\uparrow$& AbsRel$\downarrow$ & $\delta1$$\uparrow$  \\
        \midrule
        \cmark& \cmark& & &\multicolumn{2}{c}{0.911} & 0.196  & 0.727 &0.239 &0.519 &  0.126&0.840  & 0.157&0.786  \\
        \cmark& \cmark & \cmark& &\multicolumn{2}{c}{0.914} &0.196 &0.731  &0.239  &0.519& 0.126&0.839 &0.155& 0.789 \\
         & \cmark & \cmark & \cmark &\multicolumn{2}{c}{0.916} &0.197 &0.725  &0.239  &0.538& 0.127&0.839 &0.156& 0.789  \\ 
        \cmark& \cmark&\cmark & \cmark &\multicolumn{2}{c}{ \textbf{0.923}}&  0.198 & \textbf{0.727} &\textbf{ 0.227} & \textbf{0.555} &\textbf{0.126}&\textbf{0.841}&\textbf{0.153} &\textbf{0.793} \\
        \bottomrule
    \end{tabular}
    \caption{ Ablation study of different components. $\mathcal{P}$ denotes the perturbation.}
    \label{tab:ablation}
\end{table}

\textbf{Performance on KITTI-C benchmarks.}
KITTI-C~\cite{kong2023robodepth} is a widely used benchmark proposed by RoboDepth~\cite{kong2023robodepth}, which introduces synthetic perturbations to KITTI~\cite{kitti} images to simulate various challenging scenarios. We select four subsets: darkness, snow, motion and Gaussian noise, which reflect complexities in lighting and climate conditions. The experimental results, as presented in \tabref{tab:synthetic low quanlity}, demonstrate consistent improvements across multiple performance metrics.

\textbf{Performance on general benchmarks.}
In addition, comparisons are made on generalized datasets. As demonstrated in \tabref{tab:general_benchmark_results}, the performance of our model on the generalized dataset is comparable to that of the existing foundation MDE models. This finding suggests that the proposed approach can enhance the robustness of the depth estimation model without compromising its generalizability.

\textbf{Limited improvements on traditional benchmarks.}
Combining \tabref{tab:real low quanlity}, \tabref{tab:synthetic low quanlity}, and \tabref{tab:general_benchmark_results}, we observe that the improvements of \MyMthd{} over other foundation MDE models are somewhat limited. Similar phenomena can also be seen in the DepthPro~\cite{bochkovskii2024depthpro} and DepthAnything series~\cite{yang2024depth, yang2024depthv2}. 
As discussed in DepthAnything V2~\cite{yang2024depthv2}, existing benchmarks with ground truth collected in real-world environments often suffer from noise, limited diversity, and low resolution. Moreover, ground truths acquired from sensors such as LiDAR~\cite{kitti} or depth cameras~\cite{nyu} are typically sparse and lack fine-grained object details, thereby limiting the effectiveness of evaluating foundation monocular depth estimation (MDE) models.

Compared to these benchmarks, \MyMthd{} demonstrates more pronounced improvements on the more modern DA-2K-based benchmark, which provides precise depth relationships, high-resolution imagery, and extensive scene coverage. This highlights the advantage of evaluating on benchmarks with higher-quality ground truths and greater scene diversity.

\begin{figure}[t] 
  \centering
  \includegraphics[width=1.0\textwidth]{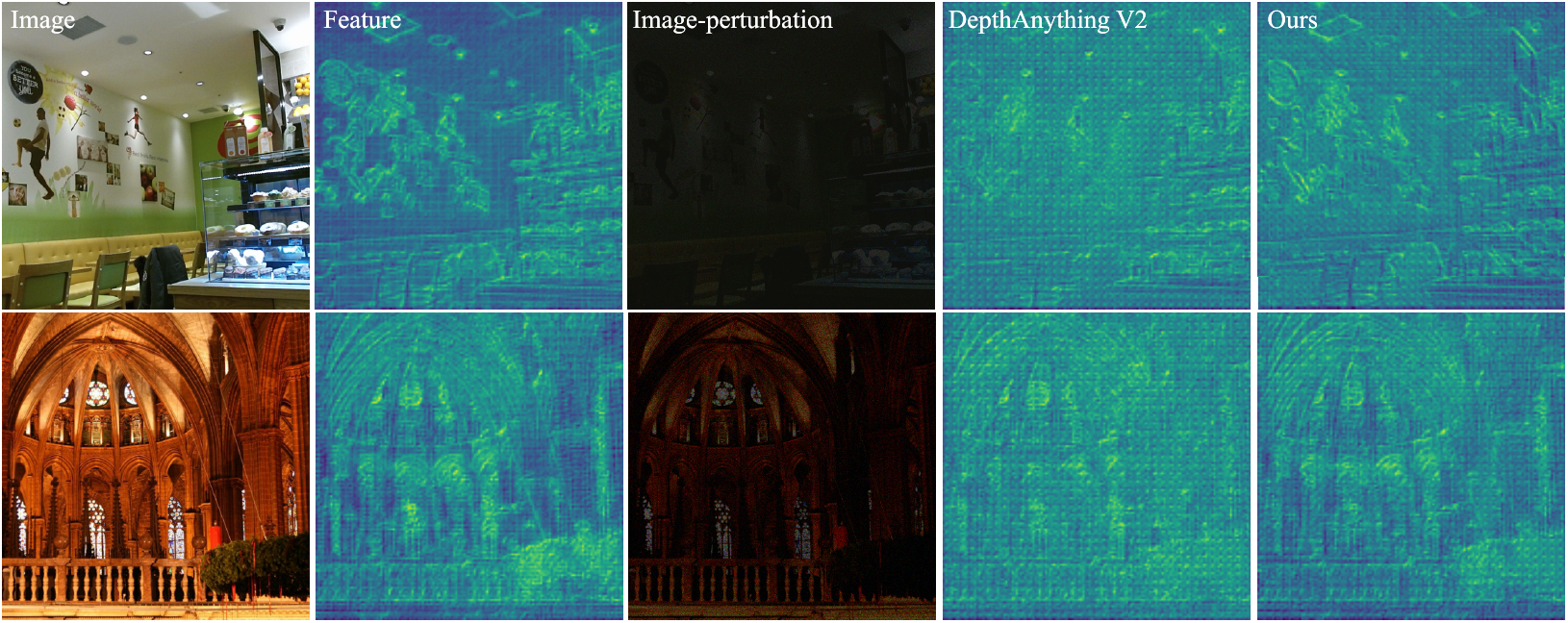} 
  \caption{
  \textbf{Visualization of feature representations. }
  `Feature' refers to the representation extracted from the original image, while the two columns on the right show the features generated from the perturbed images.
  Compared to Depth Anything V2~\cite{yang2024depthv2}, our \MyMthd{} demonstrates the ability to recover feature representations from perturbed images.
  } 
  \label{fig:feature} 
  
\end{figure}

\begin{table}[t]

      \setlength{\abovecaptionskip}{2pt}
    \begin{subtable}{.49\linewidth} 
        \centering
        \small
        \tablestyle{7pt}{1.0}
        \begin{tabular}{lcccc}
            \toprule
            \multirow{2}{*}{\textbf{Method}}  & \multicolumn{2}{c}{\textbf{NuScenes-night}} & \multicolumn{2}{c}{\textbf{Robotcar-night}} \\
            \cmidrule(lr){2-3} \cmidrule(lr){4-5}
            & AbsRel$\downarrow$ & $\delta1$$\uparrow$ & AbsRel$\downarrow$ & $\delta1$$\uparrow$ \\
            \midrule
            KD & 0.201 & 0.721 & 0.234 & 0.535 \\
            KD+Con. & 0.200& 0.724 & 0.233 & 0.538 \\
            \textbf{Con.} & \textbf{0.198} &\textbf{0.727} & \textbf{0.227} &\textbf{0.555}  \\
            
            \bottomrule
        \end{tabular}
        \caption{
        Discussion of training strategy.
        }
        \label{tab:loss mode}
    \end{subtable}
    \begin{subtable}{.5\linewidth} 
        \centering
        \small
        \tablestyle{3pt}{1.0}
        \begin{tabular}{lcccc}
            \toprule
            \multirow{2}{*}{\textbf{Loss Function}}  & \multicolumn{2}{c}{\textbf{NuScenes-night}} & \multicolumn{2}{c}{\textbf{Robotcar-night}} \\
            \cmidrule(lr){2-3} \cmidrule(lr){4-5}
            & AbsRel$\downarrow$ & $\delta1$$\uparrow$ & AbsRel$\downarrow$ & $\delta1$$\uparrow$ \\
            \midrule
            Midas ~\cite{midas}& 0.206 & 0.716 & 0.238 & 0.529 \\
            DepthAnything~\cite{yang2024depth} & 0.202 & 0.721 & 0.231 & 0.544 \\
           \textbf{Ours} & \textbf{0.198} &\textbf{0.727} & \textbf{0.227} & \textbf{0.555}\\
            \bottomrule
        \end{tabular}
        \caption{Discussion of affine-invariant loss.}
        \label{tab:loss selct}
    \end{subtable}
    \caption{Ablation studies of \MyMthd{}. `\textbf{KD}' denotes the knowledge distillation paradigm. `\textbf{Con.}' denotes the consistency regularization paradigm.}
\end{table}

\subsection{Ablation Study}

\myPara{Effectiveness of components.}
We first conduct ablation studies on different components of our \MyMthd{} to demonstrate their effectiveness in \tabref{tab:ablation}.
It can be seen that on the DA-2K dark benchmark, both the perturbation module $\mathcal{P}$ and the spatial distance constraint $\mathcal{L}_s$ lead to improvements over the consistency regularization baseline. Notably, for the setting without perturbation, we adopt the strong augmentation strategies that are commonly used in semi-supervised learning paradigms~\cite{sun2023corrmatch, yang2022revisiting}.


\myPara{Different training paradigms.}
To demonstrate the effectiveness of the proposed perturbation-based consistency regularization framework, we compare different training paradigms in \tabref{tab:loss mode}. The experimental results show that our consistency-based paradigm outperforms the knowledge distillation approach. This suggests that the non-perturbed branch of the student model possesses sufficient generalization capability to provide effective supervision and can generate higher-quality pseudo labels compared to a frozen teacher model. Furthermore, it supports the ability of \MyMthd{} to maintain competitive performance on general benchmarks, as evidenced in \tabref{tab:general_benchmark_results}.

\myPara{Different affine-invariant loss.}
As demonstrated in the  \tabref{tab:loss selct}, a comparative analysis is conducted to ascertain the discrepancy in performance between the affine-invariant loss proposed by Midas~\cite{midas}, DepthAnything~\cite{yang2024depth,yang2024depthv2}, and \MyMthd{}. The experimental findings demonstrate that the selected loss function is the most appropriate for the given conditions.

\subsection{Qualitative analysis}
In \figref{fig:feature}, we illustrate the impact of image perturbations on the feature representations of Depth Anything V2~\cite{yang2024depthv2} along with the restoration achieved by \MyMthd{}. It is evident that perturbations significantly degrade the feature quality, which may explain the poor ability of the existing foundation MDE models to perceive semantic boundaries and fine details from corrupted images. Meanwhile, \MyMthd{} expands the capability of foundation MDE models with the consistency-based framework and enhances the model’s awareness of spatial relationships throughout spatial distance constraint, thereby enabling it to recover fine-grained features from corrupted inputs.

\section{Conclusions}
In this paper, we propose \MyMthd{}, a foundation monocular depth estimation (MDE) model that extends the capabilities of existing models in complex scenarios, particularly under challenging lighting, diverse weather conditions, and distortions caused by devices, while preserving better object boundaries and details. We introduce an unsupervised perturbation-based consistency framework that enables the model to learn to handle complex scenes from a small amount of unlabeled general data. To address the inability of existing models to recover details from corrupted images, we incorporate a spatial distance constraint to enforce the geometric relative relationships between patches. Experimental results demonstrate that \MyMthd{} provides robust and fine-grained depth predictions, achieving superior performance on various benchmarks, especially on the more modern DA-2K-based benchmark, while maintaining general scene capabilities.

{
\bibliography{main}

\begin{thebibliography}{100}\itemsep=-1pt

\bibitem{bhat2021adabins}
Shariq~Farooq Bhat, Ibraheem Alhashim, and Peter Wonka.
\newblock Adabins: Depth estimation using adaptive bins.
\newblock In {\em Proceedings of the IEEE/CVF conference on computer vision and pattern recognition}, pages 4009--4018, 2021.

\bibitem{bhat2023zoedepth}
Shariq~Farooq Bhat, Reiner Birkl, Diana Wofk, Peter Wonka, and Matthias M{\"u}ller.
\newblock Zoedepth: Zero-shot transfer by combining relative and metric depth.
\newblock {\em arXiv preprint arXiv:2302.12288}, 2023.

\bibitem{bhoi2019monocular}
Amlaan Bhoi.
\newblock Monocular depth estimation: A survey.
\newblock {\em arXiv preprint arXiv:1901.09402}, 2019.

\bibitem{bian2019neurips}
Jiawang Bian, Zhichao Li, Naiyan Wang, Huangying Zhan, Chunhua Shen, Ming-Ming Cheng, and Ian Reid.
\newblock Unsupervised scale-consistent depth and ego-motion learning from monocular video.
\newblock In {\em Advances in Neural Information Processing Systems (NeurIPS)}, 2019.

\bibitem{bian2021tpami}
Jia-Wang Bian, Huangying Zhan, Naiyan Wang, Tat-Jin Chin, Chunhua Shen, and Ian Reid.
\newblock Auto-rectify network for unsupervised indoor depth estimation.
\newblock {\em IEEE Transactions on Pattern Analysis and Machine Intelligence (TPAMI)}, 2021.

\bibitem{birkl2023midas}
Reiner Birkl, Diana Wofk, and Matthias M{\"u}ller.
\newblock Midas v3.1 -- a model zoo for robust monocular relative depth estimation.
\newblock {\em arXiv preprint arXiv:2307.14460}, 2023.

\bibitem{bochkovskii2024depthpro}
Aleksei Bochkovskii, Ama{\~A}{\c{G}}l Delaunoy, Hugo Germain, Marcel Santos, Yichao Zhou, Stephan~R Richter, and Vladlen Koltun.
\newblock Depth pro: Sharp monocular metric depth in less than a second.
\newblock {\em arXiv preprint arXiv:2410.02073}, 2024.

\bibitem{sintel}
Daniel~J Butler, Jonas Wulff, Garrett~B Stanley, and Michael~J Black.
\newblock A naturalistic open source movie for optical flow evaluation.
\newblock In {\em Computer Vision--ECCV 2012: 12th European Conference on Computer Vision, Florence, Italy, October 7-13, 2012, Proceedings, Part VI 12}, pages 611--625. Springer, 2012.

\bibitem{cabon2020vkitti2}
Yohann Cabon, Naila Murray, and Martin Humenberger.
\newblock Virtual kitti 2, 2020.

\bibitem{caesar2020nuscenes}
Holger Caesar, Varun Bankiti, Alex~H Lang, Sourabh Vora, Venice~Erin Liong, Qiang Xu, Anush Krishnan, Yu Pan, Giancarlo Baldan, and Oscar Beijbom.
\newblock nuscenes: A multimodal dataset for autonomous driving.
\newblock In {\em Proceedings of the IEEE/CVF conference on computer vision and pattern recognition}, pages 11621--11631, 2020.

\bibitem{cai2024spatialbot}
Wenxiao Cai, Yaroslav Ponomarenko, Jianhao Yuan, Xiaoqi Li, Wankou Yang, Hao Dong, and Bo Zhao.
\newblock Spatialbot: Precise spatial understanding with vision language models.
\newblock {\em arXiv preprint arXiv:2406.13642}, 2024.

\bibitem{cho2021deep}
Jaehoon Cho, Dongbo Min, Youngjung Kim, and Kwanghoon Sohn.
\newblock Deep monocular depth estimation leveraging a large-scale outdoor stereo dataset.
\newblock {\em Expert Systems with Applications}, 178:114877, 2021.

\bibitem{cho2021diml}
Jaehoon Cho, Dongbo Min, Youngjung Kim, and Kwanghoon Sohn.
\newblock Diml/cvl rgb-d dataset: 2m rgb-d images of natural indoor and outdoor scenes.
\newblock {\em arXiv preprint arXiv:2110.11590}, 2021.

\bibitem{deng2024boost}
Junyuan Deng, Wei Yin, Xiaoyang Guo, Qian Zhang, Xiaotao Hu, Weiqiang Ren, Xiaoxiao Long, and Ping Tan.
\newblock Boost 3d reconstruction using diffusion-based monocular camera calibration.
\newblock {\em arXiv preprint arXiv:2411.17240}, 2024.

\bibitem{dosovitskiy2020vit}
Alexey Dosovitskiy, Lucas Beyer, Alexander Kolesnikov, Dirk Weissenborn, Xiaohua Zhai, Thomas Unterthiner, Mostafa Dehghani, Matthias Minderer, Georg Heigold, Sylvain Gelly, Jakob Uszkoreit, and Neil Houlsby.
\newblock An image is worth 16x16 words: Transformers for image recognition at scale.
\newblock {\em ICLR}, 2021.

\bibitem{eigen2014depth}
David Eigen, Christian Puhrsch, and Rob Fergus.
\newblock Depth map prediction from a single image using a multi-scale deep network.
\newblock {\em Advances in neural information processing systems}, 27, 2014.

\bibitem{everingham2010pascal}
Mark Everingham, Luc Van~Gool, Christopher~KI Williams, John Winn, and Andrew Zisserman.
\newblock The pascal visual object classes (voc) challenge.
\newblock {\em International journal of computer vision}, 88:303--338, 2010.

\bibitem{fan2025depth}
Junkai Fan, Kun Wang, Zhiqiang Yan, Xiang Chen, Shangbing Gao, Jun Li, and Jian Yang.
\newblock Depth-centric dehazing and depth-estimation from real-world hazy driving video.
\newblock In {\em Proceedings of the AAAI Conference on Artificial Intelligence}, volume~39, pages 2852--2860, 2025.

\bibitem{fournier1982computer}
Alain Fournier, Don Fussell, and Loren Carpenter.
\newblock Computer rendering of stochastic models.
\newblock {\em Communications of the ACM}, 25(6):371--384, 1982.

\bibitem{fu2018deep}
Huan Fu, Mingming Gong, Chaohui Wang, Kayhan Batmanghelich, and Dacheng Tao.
\newblock Deep ordinal regression network for monocular depth estimation.
\newblock In {\em Proceedings of the IEEE conference on computer vision and pattern recognition}, pages 2002--2011, 2018.

\bibitem{fu2024geowizard}
Xiao Fu, Wei Yin, Mu Hu, Kaixuan Wang, Yuexin Ma, Ping Tan, Shaojie Shen, Dahua Lin, and Xiaoxiao Long.
\newblock Geowizard: Unleashing the diffusion priors for 3d geometry estimation from a single image.
\newblock In {\em ECCV}, 2024.

\bibitem{gaidon2016virtual}
Adrien Gaidon, Qiao Wang, Yohann Cabon, and Eleonora Vig.
\newblock Virtual worlds as proxy for multi-object tracking analysis.
\newblock In {\em Proceedings of the IEEE conference on Computer Vision and Pattern Recognition}, pages 4340--4349, 2016.

\bibitem{r4}
Stefano Gasperini, Patrick Koch, Vinzenz Dallabetta, Nassir Navab, Benjamin Busam, and Federico Tombari.
\newblock R4dyn: Exploring radar for self-supervised monocular depth estimation of dynamic scenes.
\newblock In {\em 2021 International Conference on 3D Vision (3DV)}, pages 751--760. IEEE, 2021.

\bibitem{kitti}
Andreas Geiger, Philip Lenz, Christoph Stiller, and Raquel Urtasun.
\newblock Vision meets robotics: The kitti dataset.
\newblock {\em International Journal of Robotics Research (IJRR)}, 2013.

\bibitem{godard2017unsupervised}
Cl{\'e}ment Godard, Oisin Mac~Aodha, and Gabriel~J Brostow.
\newblock Unsupervised monocular depth estimation with left-right consistency.
\newblock In {\em Proceedings of the IEEE conference on computer vision and pattern recognition}, pages 270--279, 2017.

\bibitem{godard2019digging}
Cl{\'e}ment Godard, Oisin Mac~Aodha, Michael Firman, and Gabriel~J Brostow.
\newblock Digging into self-supervised monocular depth estimation.
\newblock In {\em Proceedings of the IEEE/CVF international conference on computer vision}, pages 3828--3838, 2019.

\bibitem{goodfellow2020generative}
Ian Goodfellow, Jean Pouget-Abadie, Mehdi Mirza, Bing Xu, David Warde-Farley, Sherjil Ozair, Aaron Courville, and Yoshua Bengio.
\newblock Generative adversarial networks.
\newblock {\em Communications of the ACM}, 63(11):139--144, 2020.

\bibitem{gui2024depthfm}
Ming Gui, Johannes Schusterbauer, Ulrich Prestel, Pingchuan Ma, Dmytro Kotovenko, Olga Grebenkova, Stefan~Andreas Baumann, Vincent~Tao Hu, and Björn Ommer.
\newblock Depthfm: Fast monocular depth estimation with flow matching, 2024.

\bibitem{guo2025murre}
Haoyu Guo, He Zhu, Sida Peng, Haotong Lin, Yunzhi Yan, Tao Xie, Wenguan Wang, Xiaowei Zhou, and Hujun Bao.
\newblock Multi-view reconstruction via sfm-guided monocular depth estimation.
\newblock In {\em CVPR}, 2025.

\bibitem{he2024lotus}
Jing He, Haodong Li, Wei Yin, Yixun Liang, Leheng Li, Kaiqiang Zhou, Hongbo Liu, Bingbing Liu, and Ying-Cong Chen.
\newblock Lotus: Diffusion-based visual foundation model for high-quality dense prediction.
\newblock {\em arXiv preprint arXiv:2409.18124}, 2024.

\bibitem{he2025distill}
Xiankang He, Dongyan Guo, Hongji Li, Ruibo Li, Ying Cui, and Chi Zhang.
\newblock Distill any depth: Distillation creates a stronger monocular depth estimator.
\newblock {\em arXiv preprint arXiv: 2502.19204}, 2025.

\bibitem{hu2024metric3dv2}
Mu Hu, Wei Yin, Chi Zhang, Zhipeng Cai, Xiaoxiao Long, Hao Chen, Kaixuan Wang, Gang Yu, Chunhua Shen, and Shaojie Shen.
\newblock Metric3d v2: A versatile monocular geometric foundation model for zero-shot metric depth and surface normal estimation.
\newblock {\em IEEE Transactions on Pattern Analysis and Machine Intelligence}, 2024.

\bibitem{hu2024drivingworld}
Xiaotao Hu, Wei Yin, Mingkai Jia, Junyuan Deng, Xiaoyang Guo, Qian Zhang, Xiaoxiao Long, and Ping Tan.
\newblock Drivingworld: Constructingworld model for autonomous driving via video gpt.
\newblock {\em arXiv preprint arXiv:2412.19505}, 2024.

\bibitem{ke2023repurposing}
Bingxin Ke, Anton Obukhov, Shengyu Huang, Nando Metzger, Rodrigo~Caye Daudt, and Konrad Schindler.
\newblock Repurposing diffusion-based image generators for monocular depth estimation.
\newblock In {\em Proceedings of the IEEE/CVF Conference on Computer Vision and Pattern Recognition (CVPR)}, 2024.

\bibitem{kerbl20233d}
Bernhard Kerbl, Georgios Kopanas, Thomas Leimk{\"u}hler, and George Drettakis.
\newblock 3d gaussian splatting for real-time radiance field rendering.
\newblock {\em ACM Trans. Graph.}, 42(4):139--1, 2023.

\bibitem{kim2017deep}
Sunok Kim, Dongbo Min, Bumsub Ham, Seungryong Kim, and Kwanghoon Sohn.
\newblock Deep stereo confidence prediction for depth estimation.
\newblock In {\em 2017 ieee international conference on image processing (icip)}, pages 992--996. IEEE, 2017.

\bibitem{kim2016structure}
Youngjung Kim, Bumsub Ham, Changjae Oh, and Kwanghoon Sohn.
\newblock Structure selective depth superresolution for rgb-d cameras.
\newblock {\em IEEE Transactions on Image Processing}, 25(11):5227--5238, 2016.

\bibitem{kim2018deep}
Youngjung Kim, Hyungjoo Jung, Dongbo Min, and Kwanghoon Sohn.
\newblock Deep monocular depth estimation via integration of global and local predictions.
\newblock {\em IEEE transactions on Image Processing}, 27(8):4131--4144, 2018.

\bibitem{kirillov2023segment}
Alexander Kirillov, Eric Mintun, Nikhila Ravi, Hanzi Mao, Chloe Rolland, Laura Gustafson, Tete Xiao, Spencer Whitehead, Alexander~C Berg, Wan-Yen Lo, et~al.
\newblock Segment anything.
\newblock In {\em Proceedings of the IEEE/CVF international conference on computer vision}, pages 4015--4026, 2023.

\bibitem{kong2023robodepth}
Lingdong Kong, Shaoyuan Xie, Hanjiang Hu, Lai~Xing Ng, Benoit~R. Cottereau, and Wei~Tsang Ooi.
\newblock Robodepth: Robust out-of-distribution depth estimation under corruptions.
\newblock In {\em Advances in Neural Information Processing Systems}, 2023.

\bibitem{li2024visual}
Jinming Li, Wanying Wang, Yaxin Peng, Chaomin Shen, Yichen Zhu, and Zhiyuan Xu.
\newblock Visual robotic manipulation with depth-aware pretraining.
\newblock In {\em 2024 IEEE International Conference on Robotics and Biomimetics (ROBIO)}, pages 843--850. IEEE, 2024.

\bibitem{li2024manipllm}
Xiaoqi Li, Mingxu Zhang, Yiran Geng, Haoran Geng, Yuxing Long, Yan Shen, Renrui Zhang, Jiaming Liu, and Hao Dong.
\newblock Manipllm: Embodied multimodal large language model for object-centric robotic manipulation.
\newblock In {\em Proceedings of the IEEE/CVF Conference on Computer Vision and Pattern Recognition}, pages 18061--18070, 2024.

\bibitem{li2023patchfusion}
Zhenyu Li, Shariq~Farooq Bhat, and Peter Wonka.
\newblock Patchfusion: An end-to-end tile-based framework for high-resolution monocular metric depth estimation.
\newblock 2024.

\bibitem{MegaDepthLi18}
Zhengqi Li and Noah Snavely.
\newblock Megadepth: Learning single-view depth prediction from internet photos.
\newblock In {\em Computer Vision and Pattern Recognition (CVPR)}, 2018.

\bibitem{li2024aodraw}
Zhong-Yu Li, Xin Jin, Boyuan Sun, Chun-Le Guo, and Ming-Ming Cheng.
\newblock Towards raw object detection in diverse conditions.
\newblock {\em arXiv preprint arXiv:2411.15678}, 2024.

\bibitem{lin2024promptda}
Haotong Lin, Sida Peng, Jingxiao Chen, Songyou Peng, Jiaming Sun, Minghuan Liu, Hujun Bao, Jiashi Feng, Xiaowei Zhou, and Bingyi Kang.
\newblock Prompting depth anything for 4k resolution accurate metric depth estimation.
\newblock 2024.

\bibitem{lin2014microsoft}
Tsung-Yi Lin, Michael Maire, Serge Belongie, James Hays, Pietro Perona, Deva Ramanan, Piotr Doll{\'a}r, and C~Lawrence Zitnick.
\newblock Microsoft coco: Common objects in context.
\newblock In {\em Computer vision--ECCV 2014: 13th European conference, zurich, Switzerland, September 6-12, 2014, proceedings, part v 13}, pages 740--755. Springer, 2014.

\bibitem{liu2010single}
Beyang Liu, Stephen Gould, and Daphne Koller.
\newblock Single image depth estimation from predicted semantic labels.
\newblock In {\em 2010 IEEE computer society conference on computer vision and pattern recognition}, pages 1253--1260. IEEE, 2010.

\bibitem{ADDS}
Lina Liu, Xibin Song, Mengmeng Wang, Yong Liu, and Liangjun Zhang.
\newblock Self-supervised monocular depth estimation for all day images using domain separation.
\newblock In {\em Proceedings of the IEEE/CVF International Conference on Computer Vision}, pages 12737--12746, 2021.

\bibitem{liu2024depthlab}
Zhiheng Liu, Ka~Leong Cheng, Qiuyu Wang, Shuzhe Wang, Hao Ouyang, Bin Tan, Kai Zhu, Yujun Shen, Qifeng Chen, and Ping Luo.
\newblock Depthlab: From partial to complete.
\newblock {\em arXiv preprint arXiv:2412.18153}, 2024.

\bibitem{loshchilov2017decoupled}
Ilya Loshchilov and Frank Hutter.
\newblock Decoupled weight decay regularization.
\newblock {\em arXiv preprint arXiv:1711.05101}, 2017.

\bibitem{robotcar}
Will Maddern, Geoffrey Pascoe, Chris Linegar, and Paul Newman.
\newblock 1 year, 1000 km: The oxford robotcar dataset.
\newblock {\em The International Journal of Robotics Research}, 36(1):3--15, 2017.

\bibitem{mao2024stealing}
Yifan Mao, Jian Liu, and Xianming Liu.
\newblock Stealing stable diffusion prior for robust monocular depth estimation.
\newblock {\em arXiv preprint arXiv:2403.05056}, 2024.

\bibitem{nyu}
Pushmeet~Kohli Nathan~Silberman, Derek~Hoiem and Rob Fergus.
\newblock Indoor segmentation and support inference from rgbd images.
\newblock In {\em ECCV}, 2012.

\bibitem{pang2024depthhelpsimprovingpretrained}
Xincheng Pang, Wenke Xia, Zhigang Wang, Bin Zhao, Di Hu, Dong Wang, and Xuelong Li.
\newblock Depth helps: Improving pre-trained rgb-based policy with depth information injection, 2024.

\bibitem{pham2024sharpdepth}
Duc-Hai Pham, Tung Do, Phong Nguyen, Binh-Son Hua, Khoi Nguyen, and Rang Nguyen.
\newblock Sharpdepth: Sharpening metric depth predictions using diffusion distillation.
\newblock {\em arXiv preprint arXiv:2411.18229}, 2024.

\bibitem{rajapaksha2024deep}
Uchitha Rajapaksha, Ferdous Sohel, Hamid Laga, Dean Diepeveen, and Mohammed Bennamoun.
\newblock Deep learning-based depth estimation methods from monocular image and videos: A comprehensive survey.
\newblock {\em ACM Computing Surveys}, 56(12):1--51, 2024.

\bibitem{midas}
Ren\'{e} Ranftl, Katrin Lasinger, David Hafner, Konrad Schindler, and Vladlen Koltun.
\newblock Towards robust monocular depth estimation: Mixing datasets for zero-shot cross-dataset transfer.
\newblock {\em IEEE Transactions on Pattern Analysis and Machine Intelligence}, 44(3), 2022.

\bibitem{Ranftl2022}
Ren\'{e} Ranftl, Katrin Lasinger, David Hafner, Konrad Schindler, and Vladlen Koltun.
\newblock Towards robust monocular depth estimation: Mixing datasets for zero-shot cross-dataset transfer.
\newblock {\em IEEE Transactions on Pattern Analysis and Machine Intelligence}, 44(3), 2022.

\bibitem{rombach2021highresolution}
Robin Rombach, Andreas Blattmann, Dominik Lorenz, Patrick Esser, and Björn Ommer.
\newblock High-resolution image synthesis with latent diffusion models, 2021.

\bibitem{saunders2023robust}
Kieran Saunders, George Vogiatzis, and Luis~J Manso.
\newblock Self-supervised monocular depth estimation: Let's talk about the weather.
\newblock In {\em Proceedings of the IEEE/CVF International Conference on Computer Vision}, pages 8907--8917, 2023.

\bibitem{saxena2005learning}
Ashutosh Saxena, Sung Chung, and Andrew Ng.
\newblock Learning depth from single monocular images.
\newblock {\em Advances in neural information processing systems}, 18, 2005.

\bibitem{saxena2023monocular}
Saurabh Saxena, Abhishek Kar, Mohammad Norouzi, and David~J Fleet.
\newblock Monocular depth estimation using diffusion models.
\newblock {\em arXiv preprint arXiv:2302.14816}, 2023.

\bibitem{eth3d}
Thomas Sch\"ops, Johannes~L. Sch\"onberger, Silvano Galliani, Torsten Sattler, Konrad Schindler, Marc Pollefeys, and Andreas Geiger.
\newblock A multi-view stereo benchmark with high-resolution images and multi-camera videos.
\newblock In {\em Conference on Computer Vision and Pattern Recognition (CVPR)}, 2017.

\bibitem{shriram2024realmdreamer}
Jaidev Shriram, Alex Trevithick, Lingjie Liu, and Ravi Ramamoorthi.
\newblock Realmdreamer: Text-driven 3d scene generation with inpainting and depth diffusion.
\newblock 2025.

\bibitem{song2025depthmaster}
Ziyang Song, Zerong Wang, Bo Li, Hao Zhang, Ruijie Zhu, Li Liu, Peng-Tao Jiang, and Tianzhu Zhang.
\newblock Depthmaster: Taming diffusion models for monocular depth estimation.
\newblock {\em arXiv preprint arXiv:2501.02576}, 2025.

\bibitem{zhu2023ecdepth}
Ziyang Song, Ruijie Zhu, Chuxin Wang, Jiacheng Deng, Jianfeng He, and Tianzhu Zhang.
\newblock Ec-depth: Exploring the consistency of self-supervised monocular depth estimation under challenging scenes.
\newblock {\em arXiv preprint arXiv:2310.08044}, 2023.

\bibitem{defeat-net}
Jaime Spencer, Richard Bowden, and Simon Hadfield.
\newblock Defeat-net: General monocular depth via simultaneous unsupervised representation learning.
\newblock In {\em Proceedings of the IEEE Conference on Computer Vision and Pattern Recognition}, 2020.

\bibitem{sun2023corrmatch}
Boyuan Sun, Yuqi Yang, Le Zhang, Ming-Ming Cheng, and Qibin Hou.
\newblock Corrmatch: Label propagation via correlation matching for semi-supervised semantic segmentation.
\newblock {\em IEEE Computer Vision and Pattern Recognition (CVPR)}, 2024.

\bibitem{sun2025llava}
Boyuan Sun, Jiaxing Zhao, Xihan Wei, and Qibin Hou.
\newblock Llava-scissor: Token compression with semantic connected components for video llms.
\newblock {\em arXiv preprint arXiv:2506.21862}, 2025.

\bibitem{sc_depthv3}
Libo Sun, Jia-Wang Bian, Huangying Zhan, Wei Yin, Ian Reid, and Chunhua Shen.
\newblock Sc-depthv3: Robust self-supervised monocular depth estimation for dynamic scenes.
\newblock {\em IEEE Transactions on Pattern Analysis and Machine Intelligence (TPAMI)}, 2023.

\bibitem{torralba2002depth}
Antonio Torralba and Aude Oliva.
\newblock Depth estimation from image structure.
\newblock {\em IEEE Transactions on pattern analysis and machine intelligence}, 24(9):1226--1238, 2002.

\bibitem{tosi2024diffusion}
Fabio Tosi, Pierluigi Zama~Ramirez, and Matteo Poggi.
\newblock Diffusion models for monocular depth estimation: Overcoming challenging conditions.
\newblock In {\em European Conference on Computer Vision (ECCV)}, 2024.

\bibitem{ADFA}
Madhu Vankadari, Sourav Garg, Anima Majumder, Swagat Kumar, and Ardhendu Behera.
\newblock Unsupervised monocular depth estimation for night-time images using adversarial domain feature adaptation.
\newblock In {\em Computer Vision--ECCV 2020: 16th European Conference, Glasgow, UK, August 23--28, 2020, Proceedings, Part XXVIII 16}, pages 443--459. Springer, 2020.

\bibitem{vankadari2022sun}
Madhu Vankadari, Stuart Golodetz, Sourav Garg, Sangyun Shin, Andrew Markham, and Niki Trigoni.
\newblock When the sun goes down: Repairing photometric losses for all-day depth estimation.
\newblock {\em arXiv preprint arXiv:2206.13850}, 2022.

\bibitem{diode_dataset}
Igor Vasiljevic, Nick Kolkin, Shanyi Zhang, Ruotian Luo, Haochen Wang, Falcon~Z. Dai, Andrea~F. Daniele, Mohammadreza Mostajabi, Steven Basart, Matthew~R. Walter, and Gregory Shakhnarovich.
\newblock {DIODE}: {A} {D}ense {I}ndoor and {O}utdoor {DE}pth {D}ataset.
\newblock {\em CoRR}, abs/1908.00463, 2019.

\bibitem{wang2025vggt}
Jianyuan Wang, Minghao Chen, Nikita Karaev, Andrea Vedaldi, Christian Rupprecht, and David Novotny.
\newblock Vggt: Visual geometry grounded transformer.
\newblock In {\em Proceedings of the IEEE/CVF Conference on Computer Vision and Pattern Recognition}, 2025.

\bibitem{wang2024weatherdepth}
Jiyuan Wang, Chunyu Lin, Lang Nie, Shujun Huang, Yao Zhao, Xing Pan, and Rui Ai.
\newblock Weatherdepth: Curriculum contrastive learning for self-supervised depth estimation under adverse weather conditions.
\newblock In {\em 2024 IEEE International Conference on Robotics and Automation (ICRA)}, pages 4976--4982. IEEE, 2024.

\bibitem{wang2024digging}
Jiyuan Wang, Chunyu Lin, Lang Nie, Kang Liao, Shuwei Shao, and Yao Zhao.
\newblock Digging into contrastive learning for robust depth estimation with diffusion models.
\newblock In {\em Proceedings of the 32nd ACM International Conference on Multimedia}, pages 4129--4137, 2024.

\bibitem{RNW}
Kun Wang, Zhenyu Zhang, Zhiqiang Yan, Xiang Li, Baobei Xu, Jun Li, and Jian Yang.
\newblock Regularizing nighttime weirdness: Efficient self-supervised monocular depth estimation in the dark.
\newblock In {\em Proceedings of the IEEE/CVF international conference on computer vision}, pages 16055--16064, 2021.

\bibitem{wang2021depth}
Li Wang, Liang Du, Xiaoqing Ye, Yanwei Fu, Guodong Guo, Xiangyang Xue, Jianfeng Feng, and Li Zhang.
\newblock Depth-conditioned dynamic message propagation for monocular 3d object detection.
\newblock In {\em Proceedings of the IEEE/CVF Conference on Computer Vision and Pattern Recognition}, pages 454--463, 2021.

\bibitem{wang2025tacodepthefficientradarcameradepth}
Yiran Wang, Jiaqi Li, Chaoyi Hong, Ruibo Li, Liusheng Sun, Xiao Song, Zhe Wang, Zhiguo Cao, and Guosheng Lin.
\newblock Tacodepth: Towards efficient radar-camera depth estimation with one-stage fusion.
\newblock In {\em CVPR}, 2025.

\bibitem{wedel2006realtime}
Andreas Wedel, Uwe Franke, Jens Klappstein, Thomas Brox, and Daniel Cremers.
\newblock Realtime depth estimation and obstacle detection from monocular video.
\newblock In {\em Joint Pattern Recognition Symposium}, pages 475--484. Springer, 2006.

\bibitem{wu2022toward}
Cho-Ying Wu, Jialiang Wang, Michael Hall, Ulrich Neumann, and Shuochen Su.
\newblock Toward practical monocular indoor depth estimation.
\newblock In {\em CVPR}, 2022.

\bibitem{Xian_2020_CVPR}
Ke Xian, Jianming Zhang, Oliver Wang, Long Mai, Zhe Lin, and Zhiguo Cao.
\newblock Structure-guided ranking loss for single image depth prediction.
\newblock In {\em The IEEE/CVF Conference on Computer Vision and Pattern Recognition (CVPR)}, June 2020.

\bibitem{xu2024depthsplat}
Haofei Xu, Songyou Peng, Fangjinhua Wang, Hermann Blum, Daniel Barath, Andreas Geiger, and Marc Pollefeys.
\newblock Depthsplat: Connecting gaussian splatting and depth.
\newblock In {\em CVPR}, 2025.

\bibitem{yan2025synthetic}
Weilong Yan, Ming Li, Haipeng Li, Shuwei Shao, and Robby~T Tan.
\newblock Synthetic-to-real self-supervised robust depth estimation via learning with motion and structure priors.
\newblock {\em arXiv preprint arXiv:2503.20211}, 2025.

\bibitem{yang2019drivingstereo}
Guorun Yang, Xiao Song, Chaoqin Huang, Zhidong Deng, Jianping Shi, and Bolei Zhou.
\newblock Drivingstereo: A large-scale dataset for stereo matching in autonomous driving scenarios.
\newblock In {\em IEEE Conference on Computer Vision and Pattern Recognition (CVPR)}, 2019.

\bibitem{yang2024depth}
Lihe Yang, Bingyi Kang, Zilong Huang, Xiaogang Xu, Jiashi Feng, and Hengshuang Zhao.
\newblock Depth anything: Unleashing the power of large-scale unlabeled data.
\newblock In {\em Proceedings of the IEEE/CVF Conference on Computer Vision and Pattern Recognition}, pages 10371--10381, 2024.

\bibitem{yang2024depthv2}
Lihe Yang, Bingyi Kang, Zilong Huang, Zhen Zhao, Xiaogang Xu, Jiashi Feng, and Hengshuang Zhao.
\newblock Depth anything v2.
\newblock {\em Advances in Neural Information Processing Systems}, 37:21875--21911, 2024.

\bibitem{yang2022revisiting}
Lihe Yang, Lei Qi, Litong Feng, Wayne Zhang, and Yinghuan Shi.
\newblock Revisiting weak-to-strong consistency in semi-supervised semantic segmentation.
\newblock In {\em CVPR}, 2023.

\bibitem{yin2024dformer}
Bowen Yin, Xuying Zhang, Zhong-Yu Li, Li Liu, Ming-Ming Cheng, and Qibin Hou.
\newblock Dformer: Rethinking rgbd representation learning for semantic segmentation.
\newblock In {\em ICLR}, 2024.

\bibitem{dformerv2}
Bo-Wen Yin, Jiao-Long Cao, Ming-Ming Cheng, and Qibin Hou.
\newblock Dformerv2: Geometry self-attention for rgbd semantic segmentation.
\newblock In {\em CVPR}, 2025.

\bibitem{yin2023metric3d}
Wei Yin, Chi Zhang, Hao Chen, Zhipeng Cai, Gang Yu, Kaixuan Wang, Xiaozhi Chen, and Chunhua Shen.
\newblock Metric3d: Towards zero-shot metric 3d prediction from a single image.
\newblock In {\em Proceedings of the IEEE/CVF International Conference on Computer Vision}, pages 9043--9053, 2023.

\bibitem{yin2022towards}
Wei Yin, Jianming Zhang, Oliver Wang, Simon Niklaus, Simon Chen, Yifan Liu, and Chunhua Shen.
\newblock Towards accurate reconstruction of 3d scene shape from a single monocular image.
\newblock {\em IEEE Transactions on Pattern Analysis and Machine Intelligence (TPAMI)}, 2022.

\bibitem{Wei2021CVPR}
Wei Yin, Jianming Zhang, Oliver Wang, Simon Niklaus, Long Mai, Simon Chen, and Chunhua Shen.
\newblock Learning to recover 3d scene shape from a single image.
\newblock In {\em Proc. IEEE Conf. Comp. Vis. Patt. Recogn. (CVPR)}, 2021.

\bibitem{msnerf_2025_TPAMI}
Ze-Xin Yin, Peng-Yi Jiao, Jiaxiong Qiu, Ming-Ming Cheng, and Bo Ren.
\newblock Ms-nerf: Multi-space neural radiance fields.
\newblock {\em IEEE Transactions on Pattern Analysis and Machine Intelligence}, pages 1--18, 2025.

\bibitem{zhang2023adding}
Lvmin Zhang, Anyi Rao, and Maneesh Agrawala.
\newblock Adding conditional control to text-to-image diffusion models, 2023.

\bibitem{ekf-imu}
Sen Zhang, Jing Zhang, and Dacheng Tao.
\newblock Towards scale-aware, robust, and generalizable unsupervised monocular depth estimation by integrating imu motion dynamics.
\newblock In {\em European Conference on Computer Vision}, pages 143--160. Springer, 2022.

\bibitem{zhang2019pattern}
Zhenyu Zhang, Zhen Cui, Chunyan Xu, Yan Yan, Nicu Sebe, and Jian Yang.
\newblock Pattern-affinitive propagation across depth, surface normal and semantic segmentation.
\newblock In {\em Proceedings of the IEEE/CVF conference on computer vision and pattern recognition}, pages 4106--4115, 2019.

\bibitem{zhao2020monocular}
Chaoqiang Zhao, Qiyu Sun, Chongzhen Zhang, Yang Tang, and Feng Qian.
\newblock Monocular depth estimation based on deep learning: An overview.
\newblock {\em Science China Technological Sciences}, 63(9):1612--1627, 2020.

\bibitem{ITDFA}
Chaoqiang Zhao, Yang Tang, and Qiyu Sun.
\newblock Unsupervised monocular depth estimation in highly complex environments.
\newblock {\em IEEE Transactions on Emerging Topics in Computational Intelligence}, 6(5):1237--1246, 2022.

\bibitem{zhao2025llava}
Jiaxing Zhao, Boyuan Sun, Xiang Chen, Xihan Wei, and Qibin Hou.
\newblock Llava-octopus: Unlocking instruction-driven adaptive projector fusion for video understanding.
\newblock {\em arXiv preprint arXiv:2501.05067}, 2025.

\bibitem{zheng2023steps}
Yupeng Zheng, Chengliang Zhong, Pengfei Li, Huan-ang Gao, Yuhang Zheng, Bu Jin, Ling Wang, Hao Zhao, Guyue Zhou, Qichao Zhang, et~al.
\newblock Steps: Joint self-supervised nighttime image enhancement and depth estimation.
\newblock In {\em 2023 IEEE International Conference on Robotics and Automation (ICRA)}, pages 4916--4923. IEEE, 2023.

\bibitem{zhou2017scene}
Bolei Zhou, Hang Zhao, Xavier Puig, Sanja Fidler, Adela Barriuso, and Antonio Torralba.
\newblock Scene parsing through ade20k dataset.
\newblock In {\em Proceedings of the IEEE conference on computer vision and pattern recognition}, pages 633--641, 2017.

\bibitem{zhou2019semantic}
Bolei Zhou, Hang Zhao, Xavier Puig, Tete Xiao, Sanja Fidler, Adela Barriuso, and Antonio Torralba.
\newblock Semantic understanding of scenes through the ade20k dataset.
\newblock {\em International Journal of Computer Vision}, 127:302--321, 2019.

\bibitem{CycleGAN2017}
Jun-Yan Zhu, Taesung Park, Phillip Isola, and Alexei~A Efros.
\newblock Unpaired image-to-image translation using cycle-consistent adversarial networks.
\newblock In {\em Computer Vision (ICCV), 2017 IEEE International Conference on}, 2017.

\end{thebibliography}
\bibliographystyle{ieee_fullname}
}

\newpage
\appendix

\section*{Appendix}

In order to provide a more complete illustration of the capabilities of DepthAnything-AC, a comprehensive appendix has been developed. This appendix includes detailed descriptions of the evaluation benchmarks, additional experiments analysis, a variety of visualizations, and the discussion among limitations and broader impacts.


\section{Detailed Evaluation Benchmarks}

\subsection{Multi-condition DA-2K benchmark}
\textbf{DA-2K}~\cite{yang2024depthv2} is a high-resolution depth estimation dataset featuring extensive and diverse scenes. Proposed by DepthAnything V2~\cite{yang2024depthv2}, it evaluates a model’s capability by determining which of two given points lies closer to the camera plane.
Beyond this modern benchmark, we further apply perturbations to the original images to simulate four common challenging conditions: dark, fog, snow, and blur. This results in an enhanced Multi-condition DA-2K benchmark, designed to evaluate the model's robustness and effectiveness under complex visual scenarios.

\subsection{Real-world complex benchmarks}
To evaluate the model's capability in real-world complex environments, we adopt several widely used benchmarks in the robust depth estimation field~\cite{zheng2023steps, zhu2023ecdepth}, including NuScenes-Night, RobotCar-Night, and DrivingStereo. These datasets present challenging conditions such as low illumination, motion blur, and diverse weather variations, thereby providing a comprehensive assessment of model robustness.

\textbf{NuScenes-night}~\cite{caesar2020nuscenes} is a nighttime subset of the NuScenes dataset, which serves as a significant benchmark in the domain of autonomous driving. Following the split used in STEPS\cite{zheng2023steps}, we evaluate various methods on 500 real-world nighttime images, offering a realistic testbed for assessing depth estimation performance under low-light conditions.

\textbf{Robotcar-night}~\cite{robotcar} is a nighttime subset of the RobotCar benchmark, which contains both daytime and nighttime data collected in autonomous driving scenarios. Following the splits of STEPS~\cite{zheng2023steps}, we use 186 real-world nighttime images for evaluation.

\textbf{DrivingStereo}~\cite{yang2019drivingstereo} is a large-scale dataset comprising 180K images collected from real-world driving scenes under diverse and challenging weather conditions. For evaluation, we follow the official subset partition and select the rain, fog, and cloud subsets, each consisting of 500 images. These subsets are used to assess the model's performance under varying climatic scenarios, including adverse weather and reduced visibility.

\subsection{Synthetic complex benchmark KITTI-C}
KITTI-C~\cite{kong2023robodepth} is an evaluation benchmark designed to assess the robustness of depth estimation models. It augments the original KITTI dataset~\cite{kitti} with synthetic corruptions to simulate a variety of challenging scenarios, including adverse weather conditions and external disturbances such as motion blur and LiDAR beam dropout. For our evaluation, we select four representative subsets: darkness, snow, motion blur, and Gaussian noise, which cover common sources of degradation in visual perception systems.

\begin{table}[t]
    \centering
    \label{tab:augmentation}
    \tablestyle{2pt}{1.0}
        \resizebox{\textwidth}{!}{
    \begin{tabular}{m{0.2\textwidth} p{0.8\textwidth}}
        \toprule
        \multicolumn{2}{c}{\textbf{Changes in light}-Apply all} \\
        \midrule
        \multirow{4}{*}{\textbf{Dark}} & Brightness is decreased nonlinearly, with brighter regions decreasing more and darker regions less. Poisson noise is added to model random photon noise due to reduced photon counts in low-light environments, and Gaussian noise is added to model the read noise of the camera sensor in low-light conditions. \\
        \midrule
        \multicolumn{2}{c}{\textbf{Blur , Weather and Color} - Apply with certain probabilities} \\
        \midrule
        \multirow{3}{*}{\textbf{Motion Blur}} & 
        Randomly select a direction angle $\theta \in [-45^\circ,45^\circ]$, set the blur kernel radius and Gaussian parameter $\sigma$ based on severity, and construct a directional kernel via 1D Gaussian sampling with small displacements along the motion direction weighted by Gaussian coefficients.
        \\
        \midrule
         \multirow{2}{*}{\textbf{Zoom Blur}} & Generate scaling layers with incremental scaling factors using cropping and scaling techniques, and synthesize the blurring effect by averaging the equal weights. \\
         \midrule
        \multirow{3}{*}{\textbf{Fog}} & 
        Using the Diamond-Square algorithm~\cite{fournier1982computer} to generate a 2D self-similar noise field, where a severity parameter controls fog density and texture. Higher severity is associated with higher noise amplitude and reduced detail retention. 
        \\
        \midrule
         \multirow{3}{*}{\textbf{Snow}}  & 
         A Gaussian random field with parameters $(\mu,\sigma)$ is generated as the snow distribution, then snowflake size and density are adjusted via nonlinear transformation. Directional motion blur ($\theta \in[-135^\circ, -45^\circ]$) simulates falling snow.
         \\
         \midrule
        \textbf{Contrast} & Using a centered scaling transform with coefficients by $[0.05, 0.4]$. \\
        \bottomrule
    \end{tabular}}
    \caption{Specific implementation of different perturbation methods}
\end{table}

\subsection{General benchmarks}
To verify that \MyMthd{} maintains its performance on general scenes after being enhanced for robustness in complex scenarios, we follow the evaluation protocol of Depth Anything V2 and select KITTI~\cite{kitti}, NYU-D~\cite{nyu}, Sintel~\cite{sintel}, DIODE~\cite{diode_dataset}, and ETH3D~\cite{eth3d} as representative general benchmarks.

\textbf{KITTI. } The KITTI~\cite{kitti} dataset is the go-to benchmark for automated driving. Created in 2012 by the Karlsruhe Institute of Technology (KIT) and Toyota Technical Institute-Central America (TTI-C), it is used to evaluate computer vision algorithms like stereo vision, optical flow, depth estimation, and 3D target detection and tracking in real road scenarios. The dataset, collected from various sensors (including grayscale/color cameras, 64-line LIDAR, and GPS/IMU systems), contains 200,000+ 3D annotated images and 39.2 km of visual range sequences, along with annotations on occlusion and truncation of vehicles, pedestrians, bicycles, and more.

\textbf{NYU-D. } The NYUDepth V2~\cite{nyu} dataset is an authoritative indoor scene understanding dataset released by New York University (NYU) in 2012. RGB-D video sequences containing 464 diverse indoor scenes were captured through the Microsoft Kinect device, providing 1,449 pairs of densely labeled RGB vs. depth image pairs, with each image labeled with object class, instance number, and depth information.

\textbf{Sintel. } The Sintel~\cite{sintel} dataset is a high-quality optical flow estimation benchmark dataset based on the open-source animated movie "Sintel" constructed by the Max-Planck-Institut (MPI) in 2012. The database contains 1,064 combinations of stereoscopic images and densely labeled ground-truth optical flow fields, covering complex motion patterns (e.g., occlusion, motion blur, and atmospheric effects) for 23 different scenes. The dataset is divided into a training set (40 sequences) and a test set (23 sequences), and is available in two versions, "Sintel Final" (with motion blur and lighting changes) and "Sintel Clean" (without effects), and we use "Sintel Final" for evaluation.

\textbf{ETH3D. } The ETH3D~\cite{eth3d} dataset is a multimodal benchmark dataset introduced by ETH Zurich, designed for 3D reconstruction, stereo vision and evaluation of multi-view stereo matching algorithms. The dataset collects high-resolution RGB images and depth information of indoor and outdoor scenes through high-precision laser scanners and various cameras (including DSLR cameras and multi-view synchronized cameras), and provides sparse parallax annotations and dense point cloud ground truths, covering a subset of high-resolution multiviews, low-resolution multiviews, and binocular configurations.

\textbf{DIODE. } The DIODE~\cite{diode_dataset} dataset is a high-precision RGB-D dataset covering both indoor and outdoor scenes.
It is acquired by a unified sensor suite (e.g., Faro scanners), providing high-resolution, dense and wide range of depth measurements and plane normal annotations. The dataset contains 8,574 indoor and 16,884 outdoor training samples, as well as 325 indoor and 446 outdoor validation samples, supporting tasks such as monocular depth estimation and 3D reconstruction. Following the Depth Anything~\cite{yang2024depth} , We use the indoor samples for validation.

\section{More Experimental Analysis}

\subsection{Perturbations}

\begin{table}[t]
    \centering
    \small
    \tablestyle{9pt}{1.0}
    \begin{tabular}{ccccccccccccc}
        \toprule
        \multirow{2}{*}{Dark} & \multirow{2}{*}{Weather} & \multirow{2}{*}{Blur} & \multirow{2}{*}{Contrast} & \multicolumn{2}{c}{\textbf{NuScenes-night}} & \multicolumn{2}{c}{\textbf{Robotcar-night}} & \multicolumn{2}{c}{\textbf{Gaussian}}\\
        \cmidrule(lr){5-6} \cmidrule(lr){7-8} \cmidrule(lr){9-10}
        & & & & AbsRel$\downarrow$ & $\delta1$$\uparrow$ & AbsRel$\downarrow$ & $\delta1$$\uparrow$ & AbsRel$\downarrow$ & $\delta1$$\uparrow$\\
        \midrule
        \cmark & & &  & 0.199 & 0.728 & 0.239 & 0.517 &0.159 &0.785\\
        \cmark& & &  & 0.199 & 0.728 & 0.237 & 0.524 &0.158 &0.791\\
        \cmark& \cmark&&  & 0.200 & 0.727 & 0.233 & 0.540 &0.154&0.792\\
        \cmark& \cmark&\cmark &  & 0.199 & 0.728 & 0.231 & 0.543&0.158 &0.791  \\
        \cmark&\cmark & \cmark& \cmark & \textbf{0.198} & \textbf{0.727} &\textbf{0.227} & \textbf{0.555} &\textbf{0.153} &\textbf{0.793}\\
        \bottomrule
    \end{tabular}
    \caption{Impact of each perturbation type. }
    \label{tab:ablation weather way}
\end{table}
\textbf{Perturbation types.}
To assess the impact of each perturbation type on model performance, we conduct ablation experiments on \MyMthd{} in \tabref{tab:ablation weather way}. Results on the diverse Robotcar-Night benchmark reveal that each type of perturbation contributes positively to the overall performance, underscoring the necessity of incorporating each perturbation strategy.

\begin{table}[t]
      \begin{subtable}{.48\linewidth}  
              \centering
        \small
        \tablestyle{6pt}{1.0}
        \begin{tabular}{ccccc}
            \toprule
            \multirow{2}{*}{Perturbation}  & \multicolumn{2}{c}{\textbf{NuScenes-night}} & \multicolumn{2}{c}{\textbf{Robotcar-night}} \\
            \cmidrule(lr){2-3} \cmidrule(lr){4-5}
            & AbsRel$\downarrow$ & $\delta1$$\uparrow$ & AbsRel$\downarrow$ & $\delta1$$\uparrow$ \\
            \midrule
            0.7+0.8 & 0.202 & 0.718 & 0.232 & 0.544 \\
            0.5+0.6 & 0.202 & 0.721 & 0.231 & 0.545 \\
            0.3+0.4 & 0.201 & 0.725 & 0.235 & 0.533 \\
            \textbf{0.1+0.2}& \textbf{0.198} & \textbf{0.727} & \textbf{0.227} & \textbf{0.555} \\
            
            \bottomrule
        \end{tabular}
        \caption{Discussion on the probability of occurrence of blur and weather perturbation.}
        \label{tab:probability}
        \end{subtable}
      \begin{subtable}{.48\linewidth}   
        \centering
        \small
        \tablestyle{6pt}{1.0}
        \begin{tabular}{ccccc}
            \toprule
            \multirow{2}{*}{Method}  & \multicolumn{2}{c}{\textbf{NuScenes-night}} & \multicolumn{2}{c}{\textbf{Robotcar-night}} \\
            \cmidrule(lr){2-3} \cmidrule(lr){4-5}
            & AbsRel$\downarrow$ & $\delta1$$\uparrow$ & AbsRel$\downarrow$ & $\delta1$$\uparrow$ \\
            \midrule
            Con. (M) & 0.198 & 0.728 & 0.237 & 0.523 \\
            Con. (E) & 0.199 & 0.728 & 0.239 & 0.518 \\
            KD (M) & 0.198 & \textbf{0.729} & 0.237 & 0.524 \\
            \textbf{KD (E)} & \textbf{0.198} & 0.727 & \textbf{0.227} &\textbf{ 0.555} \\

            \bottomrule
        \end{tabular}
        \caption{Discussion of different training paradigms of spatial distance relations.  }
        \label{tab:SDR}
        \end{subtable}

        \caption{More ablation experiments. In (a), the left side of the figure signifies the probability of blur occurrence when each image is subjected to perturbation, whilst the right side of the figure denotes the probability of weather occurrence. In (b), `\textbf{KD}' denotes the knowledge distillation paradigm. `\textbf{Con.}' denotes the consistency regularization paradigm. `\textbf{M}' refers to Manhattan distance while `\textbf{E}' refers to Euclidean distance.}
\end{table}
\textbf{Probability of different perturbations.}
In the \tabref{tab:probability}, we present the probabilities of applying different perturbations to unlabeled images. Specifically, illumination perturbation is applied to all images, while the \tabref{tab:probability} reports the probabilities used for blur and weather perturbations. Note that contrast perturbation is grouped under weather-related perturbations. As shown, the model achieves the best performance when the probability of blur augmentation is set to 0.1 and that of weather augmentation to 0.2.

\subsection{Training scheme of Spatial Distance Relation}

As demonstrated in the \tabref{tab:SDR}, we explore the impact of the spatial distance relation on the outcomes. In this analysis, M and E represent the utilization of Manhattan or Euclidean distance in the \eqnref{eq:dp} and \eqnref{eq:final}, respectively. The findings indicate that the spatial distance relation is generated using the Euclidean distance and optimized using the knowledge distillation paradigm.

\begin{table}[t]
    \centering
    \small
    \tablestyle{7pt}{1.0}
    \begin{tabular}{lcccccccccc}
        \toprule
        \multirow{2}{*}{Method}  & \multicolumn{2}{c}{\textbf{NuScenes-night}} & \multicolumn{2}{c}{\textbf{Robotcar-night}} & \multicolumn{2}{c}{\textbf{DS-cloud}}  & \multicolumn{2}{c}{\textbf{Snow}} & \multicolumn{2}{c}{\textbf{Gaussian}}\\
        \cmidrule(lr){2-3} \cmidrule(lr){4-5}  \cmidrule(lr){6-7} \cmidrule(lr){8-9} \cmidrule(lr){10-11}
        & AbsRel$\downarrow$ & $\delta1$$\uparrow$ & AbsRel$\downarrow$ & $\delta1$$\uparrow$ & AbsRel$\downarrow$ & $\delta1$$\uparrow$&  AbsRel$\downarrow$ & $\delta1$$\uparrow$&  AbsRel$\downarrow$ & $\delta1$$\uparrow$\\
        \midrule
        Unfrozen & 0.218 & 0.691 & 0.248 & 0.494 &0.151 &0.798 &0.114&0.872&0.155&0.789\\
        \textbf{Frozen} & \textbf{0.198} &\textbf{ 0.727} & \textbf{0.227} &\textbf{ 0.555 }&\textbf{ 0.149}&\textbf{0.801} &\textbf{0.114}&\textbf{0.873}&\textbf{0.153}&\textbf{0.793}\\
        
        \bottomrule
    \end{tabular}
    \caption{The importance of maintaining the frozen state of the encoder.}
    \label{tab:frozen}
\end{table}
\subsection{Encoder Frozen vs Unfrozen}
As shown in \tabref{tab:frozen}, we investigate the impact of whether to freeze the encoder parameters during training. The results clearly indicate that including the encoder parameters in the training process leads to a decline in performance during evaluation. This suggests that freezing the encoder helps preserve its pre-trained feature representations, which proves more effective for consistent and robust performance, especially in the presence of perturbations.

\begin{table}[t]
    \centering
    \small
    \tablestyle{14pt}{1.0}
    \begin{tabular}{cccccccccc}
        \toprule
        \multirow{2}{*}{ $\lambda_1$} & \multirow{2}{*}{$\lambda_2$} & \multirow{2}{*}{$\lambda_3$} & \multicolumn{2}{c}{\textbf{NuScenes-night}} & \multicolumn{2}{c}{\textbf{Robotcar-night}} & \multicolumn{2}{c}{\textbf{Gaussian}}\\
        \cmidrule(lr){4-5} \cmidrule(lr){6-7} \cmidrule(lr){8-9}
        & & & AbsRel$\downarrow$ & $\delta1$$\uparrow$ & AbsRel$\downarrow$ & $\delta1$$\uparrow$ & AbsRel$\downarrow$ & $\delta1$$\uparrow$\\
        \midrule

        0.5& 0.3& 0.2& 0.202 & 0.720 & 0.227 &0.559 &0.156&0.790\\
        0.2& 0.5& 0.3&  0.200 & 0.727 & 0.233 & 0.540 &0.153&0.792\\
        0.3& 0.2& 0.5&  0.202 & 0.721 & 0.233 & 0.539&0.155 &0.788  \\
        1.0& 1.0&1.0 &   0.200 & 0.726 & 0.230 & 0.547 &0.155&0.791\\
        \textbf{1/3}&\textbf{1/3}&\textbf{1/3}&  \textbf{0.200} & \textbf{0.727} &\textbf{0.227} & \textbf{0.555} &\textbf{0.153} &\textbf{0.793}\\
        \bottomrule
    \end{tabular}
    \caption{Discussion about different combinations of loss weights $\lambda$.}
    \label{tab:lamda}
\end{table}
\subsection{Different loss weight}
In \tabref{tab:lamda}, we conduct more ablation experiments on different loss weights $\lambda$. 
The results demonstrate that our method is relatively insensitive to variations in the loss weight configuration. Overall, setting [$\lambda_1$, $\lambda_2$, $\lambda_3$] to [1/3, 1/3, 1/3] yields the best performance across benchmarks.

\section{More Visualization}
\subsection{ Qualitative comparison between Depth Anything V2 and \MyMthd{}}
\begin{figure}[p] 
  \centering
  \includegraphics[width=0.9\textwidth,keepaspectratio]{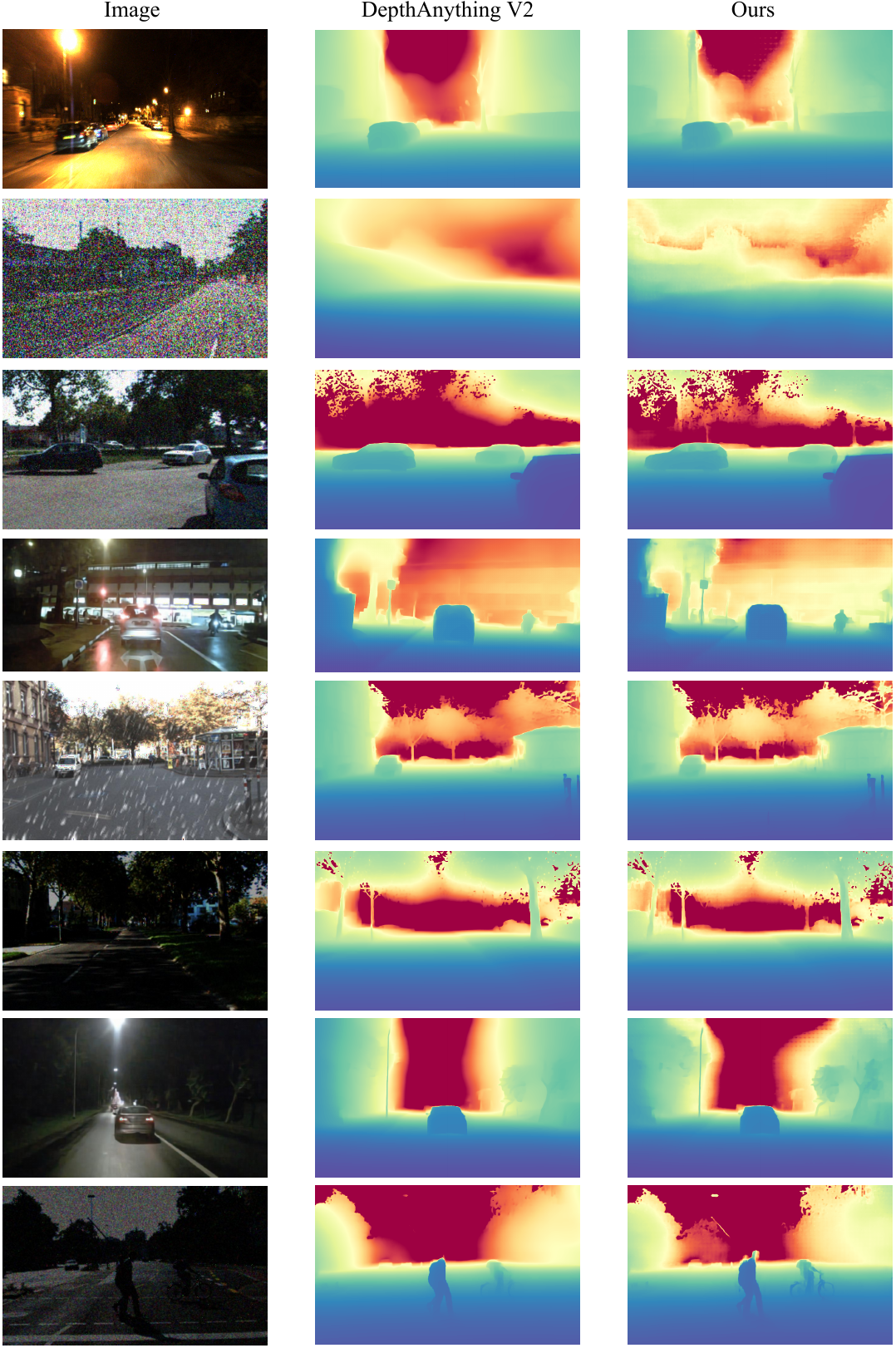} 
  \caption{Comparison between DepthAnything-V2~\cite{yang2024depthv2} and our  \MyMthd{}}
  \label{fig: more vis v2} 
\end{figure}

Please refer to \figref{fig: more vis v2}. We provide more visualization of our \MyMthd{} with comparison of Depth Anything V2~\cite{yang2024depthv2}.

\subsection{ Qualitative comparison between Depth Anything V1 and \MyMthd{}}
\begin{figure}[p] 
  \centering
  \includegraphics[width=0.9\textwidth,keepaspectratio]{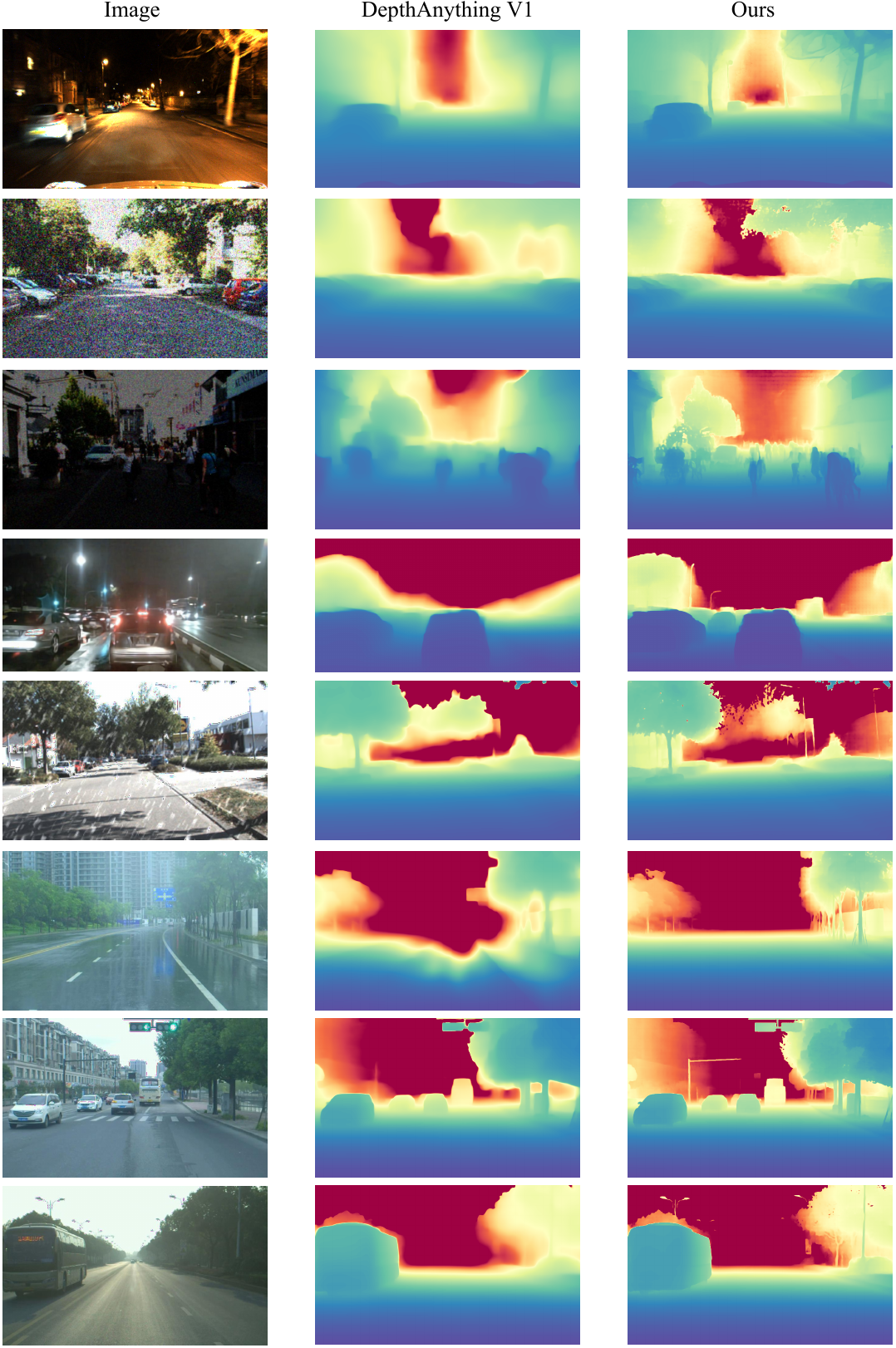} 
  \caption{Comparison between DepthAnything-V1~\cite{yang2024depth} and our \MyMthd{}}
  \label{fig: more vis v1} 
\end{figure}

Please refer to \figref{fig: more vis v1}. We provide more visualization of our \MyMthd{} with comparison of Depth Anything V1~\cite{yang2024depth}.

\subsection{ Qualitative comparison between DepthPro and \MyMthd{}}
\begin{figure}[p] 
  \centering
  \includegraphics[width=0.9\textwidth,keepaspectratio]{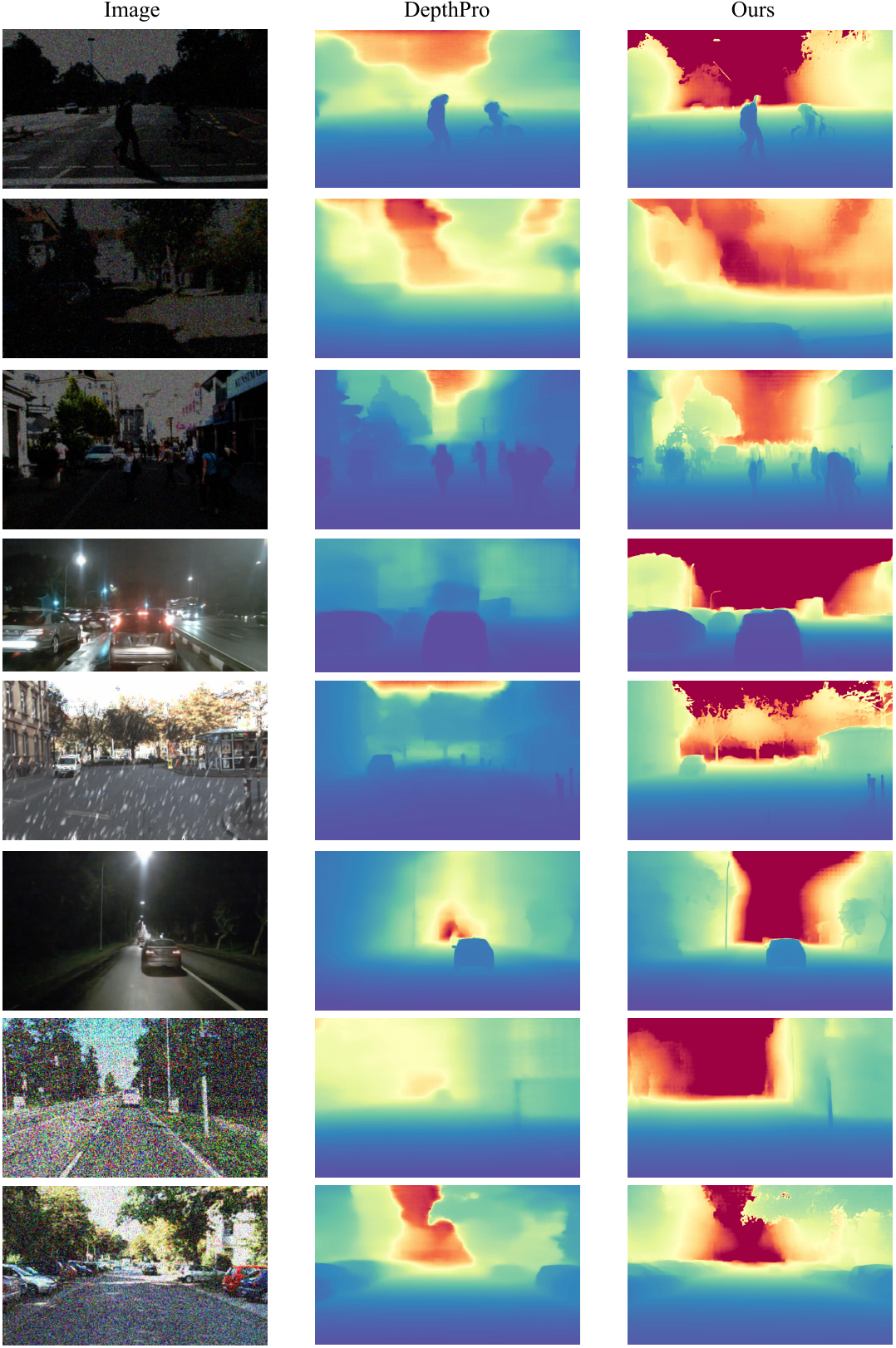} 
  \caption{Comparison between DepthPro~\cite{bochkovskii2024depthpro} and our \MyMthd{}}
  \label{fig: more vis pro} 
\end{figure}

Please refer to \figref{fig: more vis pro}. We provide more visualization of our \MyMthd{} with comparison of DepthPro~\cite{bochkovskii2024depthpro}.

\end{document}